\documentclass[10pt,journal,compsoc]{IEEEtran}
\usepackage{xcolor,soul,framed} 
\PassOptionsToPackage{hyphens}{url}\usepackage{hyperref}
\colorlet{shadecolor}{yellow}
\usepackage[pdftex]{graphicx}
\graphicspath{{../pdf/}{../jpeg/}}
\DeclareGraphicsExtensions{.pdf,.jpeg,.png}

\usepackage[cmex10]{amsmath}
\usepackage{amssymb}
\usepackage{array}
\usepackage{mdwmath}
\usepackage{mdwtab}
\usepackage{eqparbox}
\usepackage{url}

\usepackage{algorithm}
\usepackage{algorithmic}

\usepackage{multirow}
\usepackage{subfig}

\usepackage{soul}
\soulregister\cite7
\soulregister\citep7
\soulregister\ref7
\usepackage{lineno}
\newcolumntype{M}[1]{>{\centering\arraybackslash}m{#1}}
\newcolumntype{N}{@{}m{0pt}@{}}

\ifCLASSOPTIONcompsoc
  \usepackage[nocompress]{cite}
\else
  \usepackage{cite}
\fi
\ifCLASSINFOpdf
\else
\fi
\begin{document}
%
\title{AbdomenCT-1K: Is Abdominal Organ Segmentation A Solved Problem?}
%
%
%
%

\author{Jun Ma,~
       Yao Zhang, Song Gu, Cheng Zhu, Cheng Ge, Yichi Zhang, Xingle An, Congcong Wang, Qiyuan Wang, Xin Liu, Shucheng Cao, Qi Zhang, Shangqing Liu, Yunpeng Wang, Yuhui Li, Jian He,\\
       Xiaoping Yang
\IEEEcompsocitemizethanks{\IEEEcompsocthanksitem This project is supported by China's Ministry of Science and Technology (No. 2020YFA0713800) and National Natural Science Foundation of China (No. 11971229, No. 12090023). Corresponding Author: Xiaoping Yang (xpyang@nju.edu.cn).
\IEEEcompsocthanksitem Jun Ma is with Department of Mathematics, Nanjing University of Science and Technology, P.R. China. (junma@njust.edu.cn)
\IEEEcompsocthanksitem Yao Zhang is with Institute of Computing Technology, Chinese Academy of Sciences; University of Chinese Academy of Sciences, P.R. China. This work is done when Yao Zhang is an intern at AI Lab., Lenovo Research.
\IEEEcompsocthanksitem Song Gu is with School of Automation, Nanjing University of Information Science and Technology, P.R. China.
\IEEEcompsocthanksitem Cheng Zhu is with Shenzhen Haichuang Medical CO., LTD., P.R. China.
\IEEEcompsocthanksitem Cheng Ge is with Institute of Bioinformatics and Medical Engineering, Jiangsu University of Technology, P.R. China.
\IEEEcompsocthanksitem Yichi Zhang is with School of Biological Science and Medical Engineering, Beihang University, China.
\IEEEcompsocthanksitem Xingle An is with Beijing Infervision Technology CO. LTD., P.R. China.
\IEEEcompsocthanksitem Congcong Wang is with School of Computer Science and Engineering, Tianjin University of Technology, P.R. China and Department of Computer Science, Norwegian University of Science and Technology, Norway.
\IEEEcompsocthanksitem Qiyuan Wang is with School of Electronic Science and Engineering, Nanjing University, China.
\IEEEcompsocthanksitem Xin Liu is with Suzhou LungCare Medical Technology Co., Ltd, P.R. China.
\IEEEcompsocthanksitem Shucheng Cao is with Bioengineering, Biological and Environmental Science and Engineering Division, King Abdullah University of Science and Technology, Saudi Arabia
\IEEEcompsocthanksitem Qi Zhang is with Department of Computer and Information Science, Faculty of Science and Technology, University of Macau, P.R. China.
\IEEEcompsocthanksitem Shangqing Liu is with School of Biomedical Engineering, Southern Medical University, P.R. China
\IEEEcompsocthanksitem Yunpeng Wang is with Institutes of Biomedical Sciences, Fudan University, P.R. China.
\IEEEcompsocthanksitem Yuhui Li is with Computational Biology, University of Southern California, US.
\IEEEcompsocthanksitem Jian He is with  Department of Radiology, Nanjing Drum Tower Hospital, the Affiliated Hospital of Nanjing University Medical School, P.R. China.
\IEEEcompsocthanksitem Xiaoping Yang is with Department of Mathematics, Nanjing University, P.R. China.
}
}

\IEEEtitleabstractindextext{
\begin{abstract}
With the unprecedented developments in deep learning, automatic segmentation of main abdominal organs seems to be a solved problem as state-of-the-art (SOTA) methods have achieved comparable results with inter-rater variability on many benchmark datasets. However, most of the existing abdominal datasets only contain single-center, single-phase, single-vendor, or single-disease cases, and it is unclear whether the excellent performance can generalize on diverse datasets.
This paper presents a large and diverse abdominal CT organ segmentation dataset, termed AbdomenCT-1K, with more than 1000 (1K) CT scans from 12 medical centers, including multi-phase, multi-vendor, and multi-disease cases.
Furthermore, we conduct a large-scale study for liver, kidney, spleen, and pancreas segmentation and reveal the unsolved segmentation problems of the SOTA methods, such as the limited generalization ability on distinct medical centers, phases, and unseen diseases.
To advance the unsolved problems, we further build four organ segmentation benchmarks for fully supervised, semi-supervised, weakly supervised, and continual learning, which are currently challenging and active research topics.
Accordingly, we develop a simple and effective method for each benchmark, which can be used as out-of-the-box methods and strong baselines.
We believe the AbdomenCT-1K dataset will promote future in-depth research towards clinical applicable abdominal organ segmentation methods.
\end{abstract}
}

\maketitle

\IEEEdisplaynontitleabstractindextext

%
\IEEEpeerreviewmaketitle

\IEEEraisesectionheading{\section{Introduction}\label{sec:introduction}}

\begin{figure*}[!htbp]
\begin{center}
    \includegraphics[scale=0.55]{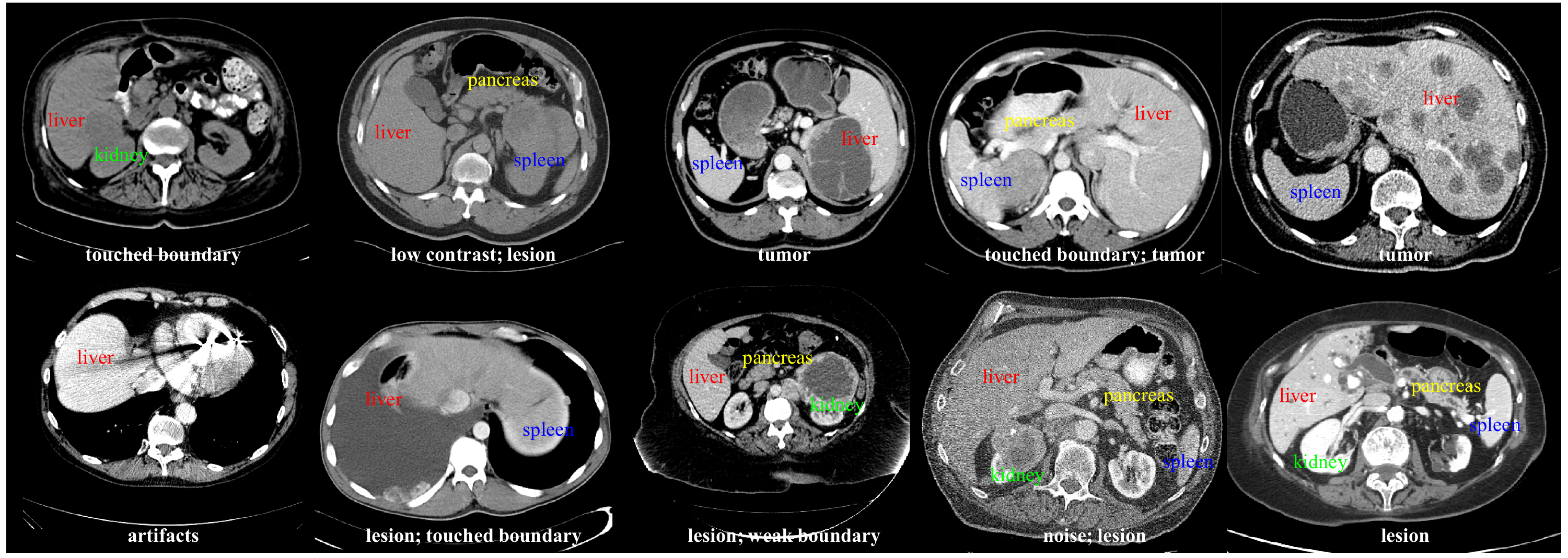}
\end{center}
\caption{Examples of abdominal organs in CT scans, including multi-center, multi-phase, multi-vendor, and multi-disease cases.}
\label{fig:hard-cases}
\end{figure*}

\IEEEPARstart{A}{bdominal} organ segmentation from medical images is an essential step for computer-assisted diagnosis, surgery navigation, visual augmentation, radiation therapy and bio-marker measurement systems~\cite{LiverSegPK09, van2011computer, sykes2014reflections, wang2019abdominal}.
In particular, computed tomography (CT) scan is one of the most commonly used  modalities for the abdominal diagnosis. It can provide structural information of multiple organs, such as liver, kidney, spleen, and pancreas, which can be used for image interpretation, surgical planning, clinical decisions, \textit{etc.}
However, the following reasons make organ segmentation a difficult task. First, the contrast of soft tissues is usually low. Second, organs may have complex morphological structures and heterogeneous lesions. Last but not least, different scanners and CT phases can lead to significant variances in organ appearances. Figure~\ref{fig:hard-cases} presents some examples of these challenging situations.

Manual contour delineation of target organs is labor-intensive and time-consuming, and also suffers from inter-~and intra- observer variability \cite{rater-variability}.
Therefore, automatic segmentation methods are highly desired in clinical studies.
In the past two decades, many abdominal segmentation methods have been proposed and massive progress has been achieved continuously in the era of deep learning. For instance, from a recently presented review work, liver segmentation can reach an accuracy of 95\% in terms of Dice similarity coefficient (DSC)~\cite{zhou2020review}. In a recent work for spleen segmentation~\cite{humpire2020fully}, 96.2\% DSC score was reported. However, most of the existing abdominal datasets only contain single-center, single-phase, single-vendor, and single-disease cases, which makes it unclear that if the performance obtained on these datasets can generalize well on more diverse datasets. Therefore, it is worth re-thinking that \textit{is abdominal organ segmentation a solved problem?}

To answer this question, in this paper, we first build a large and diverse abdominal CT organ segmentation dataset, namely AbdomenCT-1K. Then, we investigate the current limitations of the existing solutions based on the dataset. Finally, we provide four elaborately designed  benchmarks for the challenging and practical problems of abdominal organ segmentation. In the following subsections, we will summarize the limitations of the existing methods and benchmarks, and then we will briefly present the contributions of our work.

\subsection{Limitations of existing abdominal organ segmentation methods and benchmark datasets}
A clinically feasible segmentation algorithm should not only reach high accuracy, but also can generalize well on data from different sources~\cite{RSNA-Checklist, NatureMedcine-Checklist}. However, despite the encouraging progress of deep learning-based approaches and benchmarks, 
the methods and benchmarks still have some limitations that are briefly summarized as follows.

\begin{enumerate}
    \item \textbf{Lack of a large-scale and diverse dataset.}
    Evaluating the generalization ability on a large-scale and diverse dataset is highly demanded, but there exist no such kind of public dataset. As shown in Table~\ref{Tab:PubData}, most of the existing benchmark datasets either have a small number of cases or are collected from a single medical center or both.

    \item \textbf{Lack of comprehensive evaluation for the SOTA methods.}
    Most of the existing methods focus on fully supervised learning, and many of them are trained and evaluated on small publicly available datasets. It is unclear whether the proposed methods can generalize well on other testing cases, especially when the testing set is from a different medical center.

    \item \textbf{Lack of benchmarks for recently emerging annotation-efficient segmentation tasks.}
    In addition to fully supervised learning, annotation-efficient methods, such as learning with unlabelled data and weakly labelled data, have drawn many researchers' attention in both computer vision and medical image analysis communities~\cite{cheplygina2019not,tajbakhsh2020embracing,zhou2019prior,shi2020marginal}, because it is labor-intensive and time-consuming to obtain manual annotations. The availability of benchmarks plays an important role in the progress of methodology developments. For example, the SOTA performance of video segmentation has been considerably improved by the DAVIS video object segmentation benchmarks~\cite{DAVIS2020}, including semi-supervised, interactive and unsupervised tasks~\cite{caelles20192019}. However, no such kind of benchmark exists for medical image segmentation.
   Therefore, there is an urgent need to standardize the evaluation in those research fields and further boost the development of the research methodologies.

    \item \textbf{Lack of attention on organ boundary-based evaluation metrics.}
    Many of the existing benchmarks~\cite{bilic2019lits, KiTS} only use the region-based measurement (i.e., DSC) to rank segmentation methods. Boundary accuracy is also important in clinical practice~\cite{meinzer2002LiverClinical, ni2020LiverResection}, but it is insufficient to measure the boundary accuracy by DSC as demonstrated and analyzed in Figure~\ref{fig:NSD-vs-DSC}.
\end{enumerate}




\subsection{Contributions}
\begin{figure}[!htbp]
\centering
\includegraphics[scale=0.35]{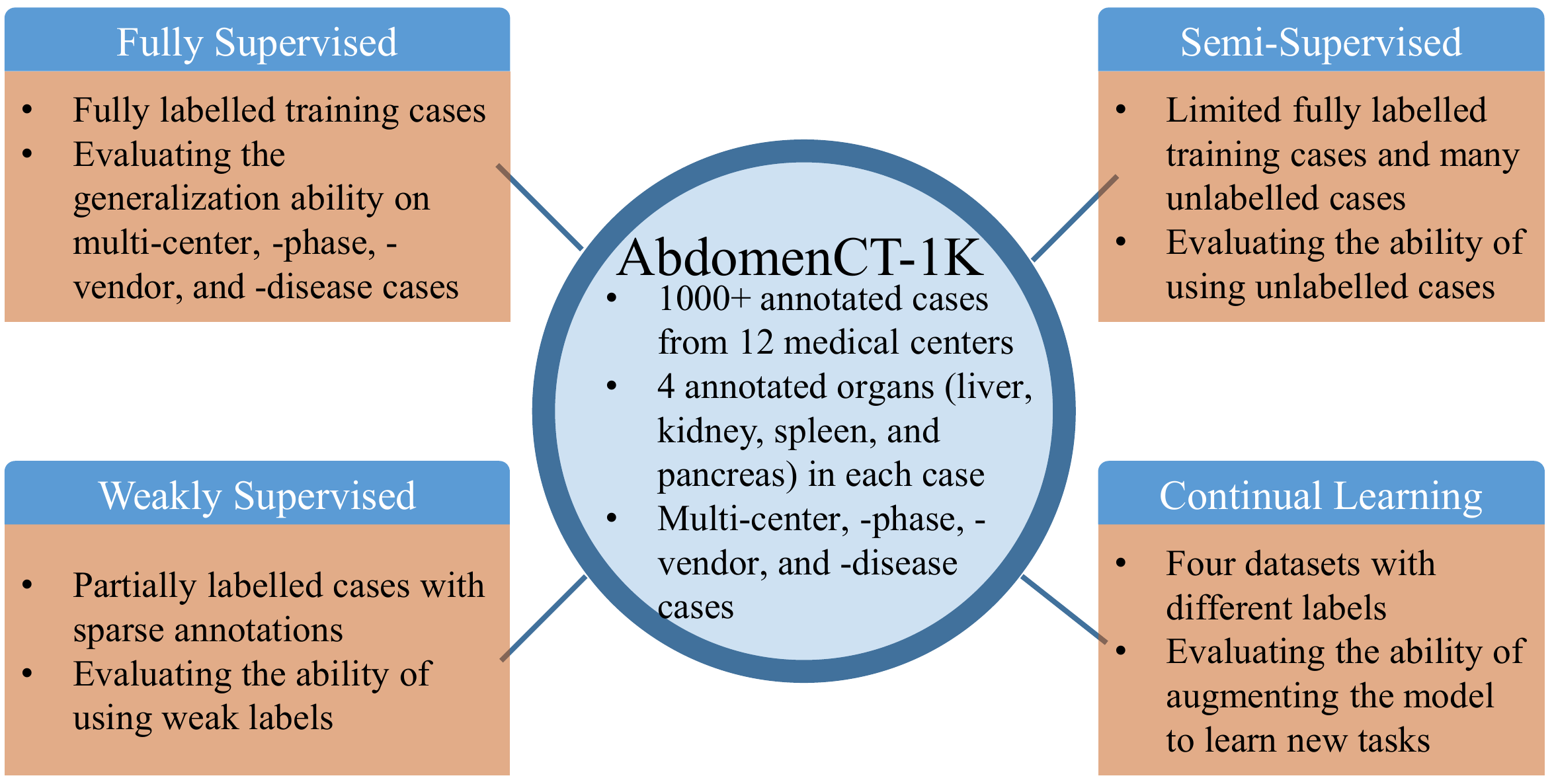}
\caption{Task overview and the associated features.}
\label{fig:task-overview}
\end{figure}

\begin{figure*}[!htbp]
\begin{center}
    \includegraphics[scale=0.5]{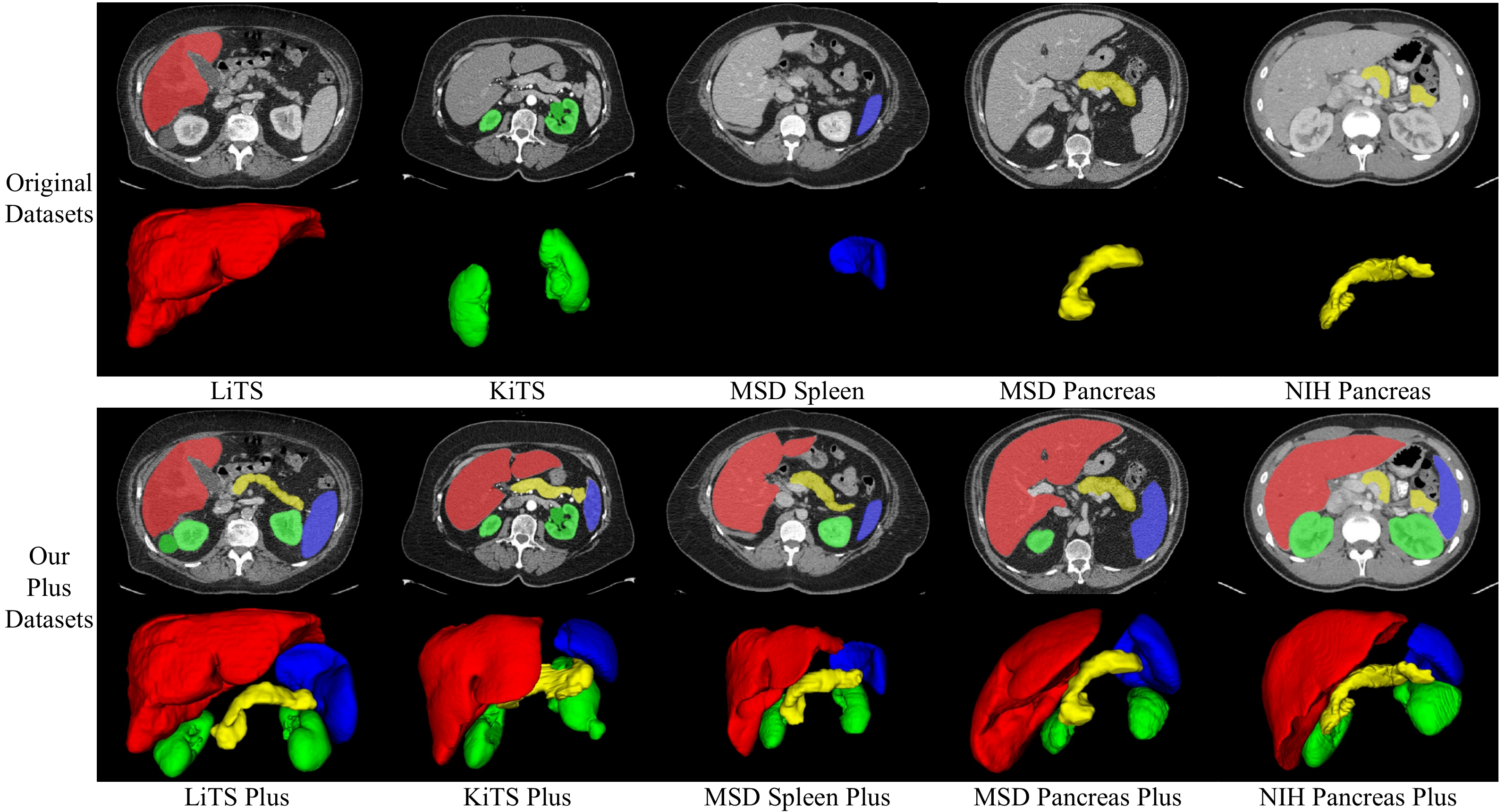}
\end{center}
\caption{Overview of the existing abdominal CT datasets and our augmented (plus) abdominal datasets. Red, green, blue, and yellow regions denote liver, kidney, spleen, and pancreas, respectively.}
\label{fig:dataoverview}
\end{figure*}

To address the above limitations, in this work, we firstly create a large-scale abdominal multi-organ CT dataset by extending the existing benchmark datasets with more organ annotations, including LiTS~\cite{bilic2019lits}, MSD~\cite{simpson2019MSD}, KiTS~\cite{KiTS},NIH-Pancreas~\cite{NIHPancreas,NIH-Pancreas2,TCIA}. Specifically, our dataset, termed AbdomenCT-1K, includes 1112 CT scans from 12 medical centers with multi-center, multi-phase, multi-vendor, and multi-disease cases. We annotate the liver, kidney, spleen, and pancreas for all cases. Figure~\ref{fig:dataoverview} and Table~\ref{Tab:PubData} illustrate the proposed AbdomenCT-1K dataset and list the main different points between our dataset and the existing abdominal organ datasets.
Then, in order to answer the question '\textit{Is abdominal organ segmentation a solved problem?}', we conduct a comprehensive study of the SOTA abdominal organ segmentation method (nnU-Net~\cite{isensee2020nnunet}) on the AbdomenCT-1K dataset for single organ and multi-organ segmentation tasks.
In addition to the widely used DSC, we add the normalized surface Dice (NSD)~\cite{nikolov2018SDice} as a boundary-based evaluation metric because the segmentation accuracy in organ boundaries is also very important in clinical practice~\cite{meinzer2002LiverClinical, ni2020LiverResection}.
Based on the results, we find that the answer is \textbf{Yes} for some ideal or easy situations, but abdominal organ segmentation is still an unsolved problem in the challenging situations, especially in the authentic clinical practice, e.g., the testing set is from a new medical center and/or contains some unseen abdominal cancer cases.
As a result, we conclude that the existing benchmarks cannot reflect the challenging cases as revealed by our large-scale study in Section~\ref{S:large-scale}. Therefore, four elaborately designed benchmarks are proposed based on AbdomenCT-1K, aiming to provide comprehensive benchmarks for fully supervised learning methods, and three annotation-efficient learning methods: semi-supervised learning, weakly supervised learning, and continual learning, which are increasingly drawing attention in the medical image analysis community.
Figure~\ref{fig:task-overview} presents an overview of our new abdominal organ benchmarks.

The main contributions of our work are summarized as follows:
\begin{enumerate}
    \item We construct, to the best of our knowledge, the up-to-date largest abdominal CT organ segmentation dataset, named AbdomenCT-1K. It contains 1112 CT scans from 12 medical centers including multi-phase, multi-vendor, and multi-disease cases. The annotations include 4446 organs (liver, kidney, spleen, and pancreas) that are significantly larger than existing abdominal organ segmentation datasets. More importantly, our dataset provides a platform for researchers to pay more attention to the generalization ability of the algorithms when developing new segmentation methodologies, which is critical for the methods to be applied in clinical practice.
    \item We conduct a large-scale study for liver, kidney, spleen, and pancreas segmentation based on the AbdomenCT-1K dataset and the SOTA method nnU-Net~\cite{isensee2020nnunet}. The extensive experiments identify some solved problems and, more importantly, reveal the unsolved problems in abdominal organ segmentation.
    \item We establish, for the first time, four new abdominal multi-organ segmentation benchmarks for fully supervised\footnote{https://abdomenct-1k-fully-supervised-learning.grand-challenge.org/}, semi-supervised\footnote{https://abdomenct-1k-semi-supervised-learning.grand-challenge.org/}, weakly supervised\footnote{https://abdomenct-1k-weaklysupervisedlearning.grand-challenge.org/}, and continual learning\footnote{https://abdomenct-1k-continual-learning.grand-challenge.org/}. These benchmarks can provide a standardized and fair evaluation of abdominal organ segmentation methods. Moreover, we also develop and provide out-of-the-box baseline solutions with the SOTA method for each task. Our dataset, code, and trained models are publicly available at \url{https://github.com/JunMa11/AbdomenCT-1K}.
\end{enumerate}

Abdominal organ segmentation in CT scans is one of the most popular segmentation tasks and there are more than 4000 teams\footnote{https://grand-challenge.org/challenges/} working on existing benchmarks.
We believe that our AbdomenCT-1K and carefully designed benchmarks can again attract the attention of the community to focus on the more challenging and practical problems in abdominal organ segmentation.

The rest of the paper is organized as follows. First, in Section~\ref{S:related_work}, the related work, including a review of abdominal organ segmentation methods and existing datasets, is presented. Then, in Section~\ref{S:abdomenct-1k}, we describe the created AbdomenCT-1K dataset. Afterwards, we conduct a comprehensive study for abdominal organ segmentation with the SOTA method nnU-Net~\cite{isensee2020nnunet} in Section~\ref{S:large-scale}, where the solved and unsolved problems for abdominal organ segmentation are also presented. Next, in order to address these unsolved problems, we set up four new benchmarks in Section~\ref{S:benchmark}, including fully supervised, semi-supervised, weakly supervised, and continual learning of abdominal organ segmentation, respectively. Finally, in Section~\ref{S:conc}, the conclusions are drawn.

\section{Related Work}
\label{S:related_work}
\subsection{Abdominal organ segmentation methods}

From the perspective of methodology, abdominal organ segmentation methods can be classified into classical model-based approaches and modern learning-based approaches.

\textbf{Model-based methods} usually formulate the image segmentation as an energy functional minimization problem or explicitly match a shape template or atlas to a new image, such as variational models~\cite{kass1988snakes}, statistical shape models~\cite{cootes1995ASM}, and atlas-based methods~\cite{iglesias2015multi}.
Level set methods or active contour models are one of the most popular variational models. They provide a natural way to drive the curves to delineate the structure of interest~\cite{li2015automatic, HuPeijun-MP, CV-SegLiver}.
Different from the level set methods, statistical shape models, such as the well-known active shape model, represent the shape of an object by a set of boundary points that are constrained by the point distribution model. Then, the model iteratively deforms the points to fit to the object in a new image~\cite{LiverSegPK09, LiverSeg-SSM}.
Atlas-based methods usually construct one or multiple organ atlas with annotated cases. Then, label fusion is used to propagate the atlas annotations to a target image via registration between the atlas image and the target image~\cite{zhou2005-atlas, okada2015abdominal, xu2015efficient}. 
Although these model-based methods have transparent principles and well-defined formulations, they usually fail to segment the organs with weak boundaries and low contrasts. Besides, the computational cost is usually high, especially for 3D CT scans.

\textbf{Learning-based methods} usually extract discriminative features from annotated CT scans to distinguish target organs and other tissues.
Since 2015, deep convolutional neural network (CNN)-based methods~\cite{ronneberger20152DUNet}, which neither rely on hand crafted features nor rely on anatomical correspondences, have been successfully introduced into abdominal organ segmentation and reach SOTA performances~\cite{ACDC-TMI, BRATS-Review}.
These approaches can be briefly classified into well-known supervised learning methods and recently emerging annotation-efficient deep learning approaches. 
In the following paragraphs, we will introduce the two categories respectively.

One group of the supervised organ segmentation methods is single organ segmentation. For example, Seo \textit{et al.} proposed a modified U-Net~\cite{seo2019modified} to segment liver and liver tumors. In~\cite{MICCAI2019-shape-prior-liver-seg}, a shape-aware method, which incorporated prior knowledge of the target organ shape into a CNN backbone, was proposed and achieved encouraging performance on liver segmentation task. While U-Net is a welcomed network structure, other backbone designs are also proposed for abdominal organ segmentation, such as progressive holistically-nested network (PHNN)~\cite{DLMI2017-PHNN,MICCAI2017-PHNN} and progressive semantically-nested networks (PSNNs) \cite{mia2021-deeptarget}. In~\cite{zhou2017fixed}, CNNs were employed to segment the pancreas. Pancreas segmentation was treated as a more challenging task compared to the liver and the kidney segmentation. Therefore, two-stage cascaded approaches were proposed~\cite{mia2018-spatial-aggre-pancreas-seg,roth2016spatial,xue2019cascaded}, where pancreas was located first, then a new network was employed to refine the segmentation. Moreover, in~\cite{TMI-LGAC}, a level set regression network was developed to obtain more accurate segmentation in pancreas boundaries. Instead of designing network structures empirically, Neural Architecture Search (NAS) technique was also introduced into organ segmentation~\cite{3DV2019-v-nas,2020cvpr-organ-at-risk-nas,2021cvpr-nas-3d-medical} by designing efficient differentiable neural architecture search strategies.

The other group of the supervised organ segmentation methods is multi-organ segmentation~\cite{roth2018application, zhou2019prior, gibson2018automatic, wang2019abdominal}, where multiple
organs are segmented simultaneously. Fully convolutional networks (FCN)-based methods have been widely applied to multi-organ segmentation. Early works include applying FCN alone~\cite{zhou2016three, gibson2018automatic} and the combinations of FCN with pre- or/and post-processing~\cite{larsson2018robust,hu2017automatic}.
However, compared to the single organ segmentation task, multi-organ segmentation is more challenging.
As shown in Figure~\ref{fig:hard-cases}, the weak boundaries between organs on CT scans and the variations of the size of different organs, make the multi-organ segmentation task harder~\cite{wang2019abdominal}.
In order to address the difficulties, cascaded networks were employed to organ segmentation. In~\cite{wang2019abdominal}, a two-stage segmentation method was proposed. An organ segmentation probability map was first computed in the first stage and was combined with the original input images for the second stage. The segmentation probability map can provide  spatial attention to the second stage, thus can enhance the target organs' discriminative information in the second stage. Other
similar strategies were proposed~\cite{roth2018application,roth2018multi}, where the first stage networks played different roles. For example, in~\cite{roth2018application}, a candidate region was generated and sent to the second stage. In~\cite{roth2018multi}, low resolution segmentation maps were extracted from the first stage.
Moreover, in~\cite{zhang2020block}, Zhang \textit{et al.} argued that the features from each intermediate layer of the first stage network can provide useful information for the second stage. Therefore, a block level skip connections (BLSC) across cascaded V-Net~\cite{milletari2016v} was proposed and showed improved performance.
In order to reduce the choices of the number of architecture layers, kernel sizes, \textit{etc.}, in~\cite{heinrich2019obelisk}, trainable 3D convolutional kernel with learnable filter coefficients and spatial offsets was presented and show its benefits to  capture large spatial context as well as the design of networks.
Noticeably, in~\cite{isensee2020nnunet}, nnU-Net, a U-Net~\cite{ronneberger20152DUNet}-based segmentation framework, was proposed and achieved state-of-the-art performances on both single organ and multi-organ segmentation tasks, including liver, kidney, pancreas, and spleen.

Recently, \textbf{annotation-efficient methods}, such as semi-supervised learning, weakly supervised learning, and continual learning, have received great attention in both computer vision and medical image analysis communities~\cite{cheplygina2019not,tajbakhsh2020embracing, zhou2020review}.
This is because fully annotated multi-organ datasets require great efforts of abdominal experts and are very expensive to obtain. Therefore, beyond fully supervised abdominal organ segmentation, some recent studies focus on learning with partially labelled organs.

\textit{Semi-supervised learning} aims to combine a small amount of labelled data with a large amount of unlabelled data, which is an effective way to explore knowledge from the unlabelled data.
It is a promising and active research direction in machine learning~\cite{van2020survey} as well as medical image analysis~\cite{cheplygina2019not}. Among the semi-supervised approaches, pseudo label-based methods are regarded as simple and efficient solutions~\cite{lee2013pseudo,iscen2019label}.
In~\cite{zhou2019semi}, a pseudo label-based semi-supervised multi-organ segmentation method was presented. A teacher model was first trained in a fully supervised way on the source dataset.
Then pseudo labels on the unlabelled dataset were computed by the trained model. Finally, a student model was trained on the combination of both the labelled and unlabelled data. Besides, other strategies are also explored. For example, in~\cite{lee2020semi}, in addition to the Dice loss computed from labelled data, a quality assurance-based discriminator module was proposed to supervise the learning on the unlabelled data. In~\cite{ECCV2020-semi-liver}, a co-training strategy was proposed to explore unlabelled data.  The proposed framework, trained on a small single phase dataset, can adapt to unlabelled multi-center and multi-phase clinical data. Moreover, an uncertainty-aware multi-view co-training (UMCT) approach was proposed in~\cite{mia2020-semi-uncertainty}, which achieves superior performance on multi-organ and pancreas datasets.

\textit{Weakly supervised learning} is to explore the use of weak annotations, such as slice-level annotations, sparse annotations, and noisy annotations~\cite{tajbakhsh2020embracing}.
For organ segmentation, in~\cite{kanavati2017joint},  a classification forest-based weakly supervised organ segmentation method was proposed for livers, spleens and kidneys, where the labels are scribbles on organs. Besides, image-level labels-based pancreas segmentation was explored in~\cite{zeng2019weakly}. Although there are limited studies related to weakly supervised learning for abdomen organ segmentation, considerable research has been done in the computer vision community for image segmentation for different weak annotations, such as bounding boxes~\cite{song2019box}, points~\cite{bearman2016s,qian2019weakly}, scribbles~\cite{lin2016scribblesup,ji2019scribble}, image-level labels~~\cite{pathak2015constrained,liu2020leveraging,wang2020weakly}.

\textit{Continual learning } is to learn new tasks without forgetting the learned tasks, which is also named as life long learning, incremental learning or sequential learning. Though deep learning methods obtain SOTA performance in many applications, neural networks suffer from catastrophic forgetting or interference~\cite{mccloskey1989catastrophic,Goodfellow14anempirical, pfulb2019comprehensive}. The learned knowledge of a model can be interfered with the new information which we train the model with. As a result, the performance of the old task could decrease. Therefore, continual learning has attracted growing attention in the past years~\cite{continualReview19}, such as object recognition~\cite{lomonaco2017core50, camoriano2017incremental} and classification~\cite{de2019continual}. Besides, tailored datasets and benchmarks for continual learning have been also proposed in the computer vision community, e.g. the object recognition dataset and benchmark CORe50~\cite{lomonaco2017core50}, iCubWorld datasets\footnote{https://robotology.github.io/iCubWorld/\#publications}, and the CVPR2020 CLVision  challange\footnote{https://sites.google.com/view/clvision2020/challenge}.
However, to the best of our knowledge, there is no continual learning work for abdominal organ segmentation. Therefore, applying this new emerging technique to tackle organ segmentation tasks is still in demand.




\begin{table*}[!htbp]
\centering
\caption{Overview of the popular  abdominal CT benchmark datasets. ``Tr/Ts'' denotes training/testing set.}
\label{Tab:PubData}
\begin{tabular}{lcccc}
\hline
Dataset Name (abbr.)                                                                            & Target          & \# of Tr/Ts & \# of Centers & Source and Year                                                                                     \\
\hline
\begin{tabular}[c]{@{}l@{}}Multi-atlas Labelling \\Beyond the Cranial Vault (BTCV)~\footnotemark[8]~\cite{BTCA2015} \end{tabular} & 13 organs          & 30/20       & 1             & MICCAI 2015                                                                                               \\
\hline
NIH Pancreas~\footnotemark[9]~\cite{NIHPancreas, NIH-Pancreas2,TCIA}                                                                                    & Pancreas           & 80          & 1             & \begin{tabular}[c]{@{}c@{}}The Cancer Imaging \\ Archive 2015 \end{tabular}  \\
\hline
\begin{tabular}[c]{@{}l@{}} VISCERAL Anatomy Benchmark~\footnotemark[10]~\cite{jimenez2016cloud} \end{tabular}           & 20 anatomical structures   & 80/40      & 1             & ISBI and ECIR 2015                                                                                               \\
\hline
\begin{tabular}[c]{@{}l@{}}Liver Tumor Segmentation \\Benchmark (LiTS)~\footnotemark[11]~\cite{bilic2019lits} \end{tabular}            & Liver and tumor    & 131/70      & 7             & ISBI and MICCAI 2017                                                                                      \\
\hline
\begin{tabular}[c]{@{}l@{}}Medical Segmentation Decathlon \\(MSD) Pancreas~\footnotemark[12]~\cite{simpson2019MSD} \end{tabular}        & Pancreas and tumor & 281/139     & 1             & MICCAI 2018                                                                                               \\
\hline
\begin{tabular}[c]{@{}l@{}}Medical Segmentation Decathlon \\(MSD) Spleen~\footnotemark[13]~\cite{simpson2019MSD} \end{tabular}          & Spleen             & 41/20       & 1             & MICCAI 2018                                                                                               \\
\hline
\begin{tabular}[c]{@{}l@{}}Multi-organ Abdominal CT \\ Reference Standard Segmentation~\footnotemark[14]~\cite{gibson2018automatic} \end{tabular}           & 8 organs   & 90      & 2             & Zenodo 2018                                                                                              \\
\hline
\begin{tabular}[c]{@{}l@{}}Combined Healthy Abdominal \\Organ Segmentation (CHAOS)~\footnotemark[15]~\cite{kavur2020chaos} \end{tabular} & Liver              & 20/20       & 1             & ISBI 2019                                                                                                 \\
\hline
\begin{tabular}[c]{@{}l@{}}Kidney Tumor Segmentation \\Benchmark (KiTS)~\footnotemark[16]~\cite{KiTSDataset} \end{tabular}           & Kidney and tumor   & 210/90      & 1             & MICCAI 2019                                                                                               \\
\hline
\begin{tabular}[c]{@{}l@{}}CT-ORG~\footnotemark[17]~\cite{rister2020ct} \end{tabular}           &  liver, lungs, bladder, kidney, bones and brain   & 119/21      & 8             & \begin{tabular}[c]{@{}c@{}} The Cancer Imaging \\ Archive 2020 \end{tabular}                                                                                              \\
\hline
\textbf{AbdomenCT-1K (ours)} & Liver, kidney, spleen, and pancreas & \textbf{1112} & \textbf{12} & 2021 \\ \hline
\end{tabular}
\end{table*}

\footnotetext[8]{https://www.synapse.org/\#!Synapse:syn3193805/wiki/89480}
\footnotetext[9]{https://wiki.cancerimagingarchive.net/display/Public/Pancreas-CT}
\footnotetext[10]{http://www.visceral.eu/benchmarks/}
\footnotetext[11]{https://competitions.codalab.org/competitions/15595}
\footnotetext[12]{http://medicaldecathlon.com/}
\footnotetext[13]{http://medicaldecathlon.com/}
\footnotetext[14]{https://zenodo.org/record/1169361\#.YMRb9NUza70}
\footnotetext[15]{https://chaos.grand-challenge.org/}
\footnotetext[16]{https://kits19.grand-challenge.org/}
\footnotetext[17]{https://wiki.cancerimagingarchive.net/display/Public/CT-ORG\%3A+CT+volumes+with+multiple+organ+segmentations}
\subsection{Existing abdominal CT organ segmentation benchmark datasets}
\label{SS:existing_dataset}
In addition to the promising progress in abdominal organ segmentation methodologies, segmentation benchmark datasets are also evolved, where the datasets contain more and more annotated cases for developing and evaluating segmentation methods.
Table \ref{Tab:PubData} summarizes the popular abdominal organ CT segmentation benchmark datasets since 2010, which will be briefly presented in the following paragraphs.

BTCV (Beyond The Cranial Vault)~\cite{BTCA2015} benchmark dataset consists of 50 abdominal CT scans acquired at the Vanderbilt University Medical Center from metastatic liver cancer patients or post-operative ventral hernia patients. 
This benchmark aims to segment 13 organs, including spleen, right kidney, left kidney, gallbladder, esophagus, liver, stomach, aorta, inferior vena cava, portal vein and splenic vein, pancreas, right adrenal gland, and left adrenal gland. The organs were manually labelled by two experienced undergraduate students, and verified by a radiologist.

NIH Pancreas dataset~\cite{NIHPancreas,NIH-Pancreas2,TCIA}, from US National Institutes of Health (NIH) Clinical Center, consists of 80 abdominal contrast enhanced 3D CT images. The CT scans have resolutions of 512$\times$512 pixels with varying pixel sizes and slice thickness between 1.5$-$2.5 mm.  Among these cases, seventeen subjects are healthy kidney donors scanned prior to nephrectomy. The remaining 65 patients were selected by a radiologist from patients who neither had major abdominal pathologies nor pancreatic cancer lesions. The pancreas was manually labelled slice-by-slice by a medical student and then verified/modified by an experienced radiologist.

VISCERAL Anatomy Benchmark~\cite{jimenez2016cloud} consists of 120 CT and MR patient volumes.
Volumes from 4 different imaging modalities and field-of-views compose the training set. Each group contains 20 volumes, which adds up to 80 volumes in the training set.
In each volume, 20 abdominal structures were manually annotated to build a standard Gold Corpus containing a total of 1295 structures and 1760 landmarks.

LiTS (Liver Tumor Segmentation) dataset~\cite{bilic2019lits} includes 131 training CT cases with liver and liver tumor annotations and 70 testing cases with hidden annotations. The images are provided with an in-plane resolution of 0.5 to 1.0 mm, and slice thickness of 0.45 to 6.0 mm. The cases are collected from 7 medical centers and the corresponding patients have a variety of primary cancers, including hepatocellular carcinoma, as well as metastatic liver disease derived from colorectal, breast, and lung primary cancers. Annotations of the liver and tumors were performed by radiologists.

MSD (Medical Segmentation Decathlon) pancreas dataset~\cite{simpson2019MSD} consists of 281 training cases with pancreas and tumor annotations and 139 testing cases with hidden annotations. The dataset is provided by Memorial Sloan Kettering Cancer Center (New York, USA).
The patients in this dataset underwent resection of pancreatic masses, including intraductal mucinous neoplasms, pancreatic neuroendocrine tumors, or pancreatic ductal adenocarcinoma.
The pancreatic parenchyma and pancreatic mass (cyst or tumor) were manually annotated in each slice by an expert abdominal radiologist.

MSD Spleen dataset~\cite{simpson2019MSD} includes 41 training cases with spleen annotations and 20 testing cases without annotations, which are also provided by Memorial Sloan Kettering Cancer Center (New York, USA).
The patients in this dataset underwent chemotherapy treatment for liver metastases.
The spleen was semi-automatically segmented using a level-set-based method and then manually adjusted by an expert abdominal radiologist.

Multi-organ Abdominal CT Reference Standard Segmentations~\cite{gibson2018automatic} is composed of 90 abdominal CT images and corresponding reference standard segmentations of 8 organs. The CT images are from the Cancer Imaging Archive (TCIA) Pancreas-CT dataset with pancreas segmentations and the Beyond the Cranial Vault (BTCV) challenge with segmentations of all organs except duodenum. The unsegmented organs were manually labelled by an imaging research fellow under the supervision of a board-certified radiologist.

CHAOS (Combined Healthy Abdominal Organ Segmentation) dataset~\cite{kavur2020chaos} consists of 20 training cases with liver annotations and 20 testing cases with hidden annotations, which are provided by Dokuz Eylul University (DEU) hospital (İzmir, Turkey). Different from the other datasets, all the 40 liver CT cases are from the healthy population.

KiTS (Kidney Tumor Segmentation) dataset~\cite{KiTSDataset} includes 210 training cases with kidney and kidney tumor annotations and 90 testing cases with hidden annotations, which are provided by the University of Minnesota Medical Center (Minnesota, USA).
The patients in this dataset underwent partial or radical nephrectomy for one or more kidney tumors.
The kidney and tumor annotations were provided by medical students under the supervision of a clinical chair.

CT-ORG~\cite{rister2020ct} is a diverse dataset of 140 CT images containing 6 organ classes, where 131 are dedicated CT and 9 are the CT component from PET-CT exams. These CT images are from 8 different medical centers. Patients were included based on the presence of lesions in one or more of the labelled organs. Most 
of the images exhibit liver lesions, both benign and malignant.



\section{AbdomenCT-1K dataset}
\label{S:abdomenct-1k}
\subsection{Dataset motivation and details}
\label{ss:dataset-motivation}
Most existing abdominal organ segmentation datasets have limitations in diversity and scale. In this paper, we present a large-scale dataset that is closer to real-world applications and has more diverse abdominal CT cases.
In particular, we focus on multi-organ segmentation, including liver, kidney, spleen, and pancreas.
To include more diverse cases, our dataset, namely AbdomenCT-1K, consists of 1112 3D CT scans from five existing datasets: LiTS (201 cases)~\cite{bilic2019lits}, KiTS (300 cases)~\cite{KiTS}, MSD Spleen (61 cases) and Pancreas (420 cases)~\cite{simpson2019MSD}, NIH Pancreas (80 cases)~\cite{NIHPancreas,NIH-Pancreas2,TCIA}, and a new dataset from Nanjing University (50 cases). The 50 CT scans in the Nanjing University dataset are from 20 patients with pancreas cancer, 20 patients with colon cancer, and 10 patients with liver cancer.  The number of plain phase, artery phase, and portal phase scans are 18, 18, and 14 respectively. The CT scans have resolutions of 512$\times$512 pixels with varying pixel sizes and slice thicknesses between 1.25-5 mm, acquired on GE multi-detector spiral CT.
The licenses of NIH Pancreas and KiTS dataset are Creative Commons license CC-BY and CC-BY-NC-SA 4.0, respectively. LiTS, MSD Pancreas, and MSD Spleen datasets are Creative Commons license CC-BY-SA 4.0. Under these licenses, we are allowed to modify the datasets and share or redistribute them in any format.

\begin{figure}[htbp]
	\begin{center}
    	\subfloat{\includegraphics[scale=0.38]{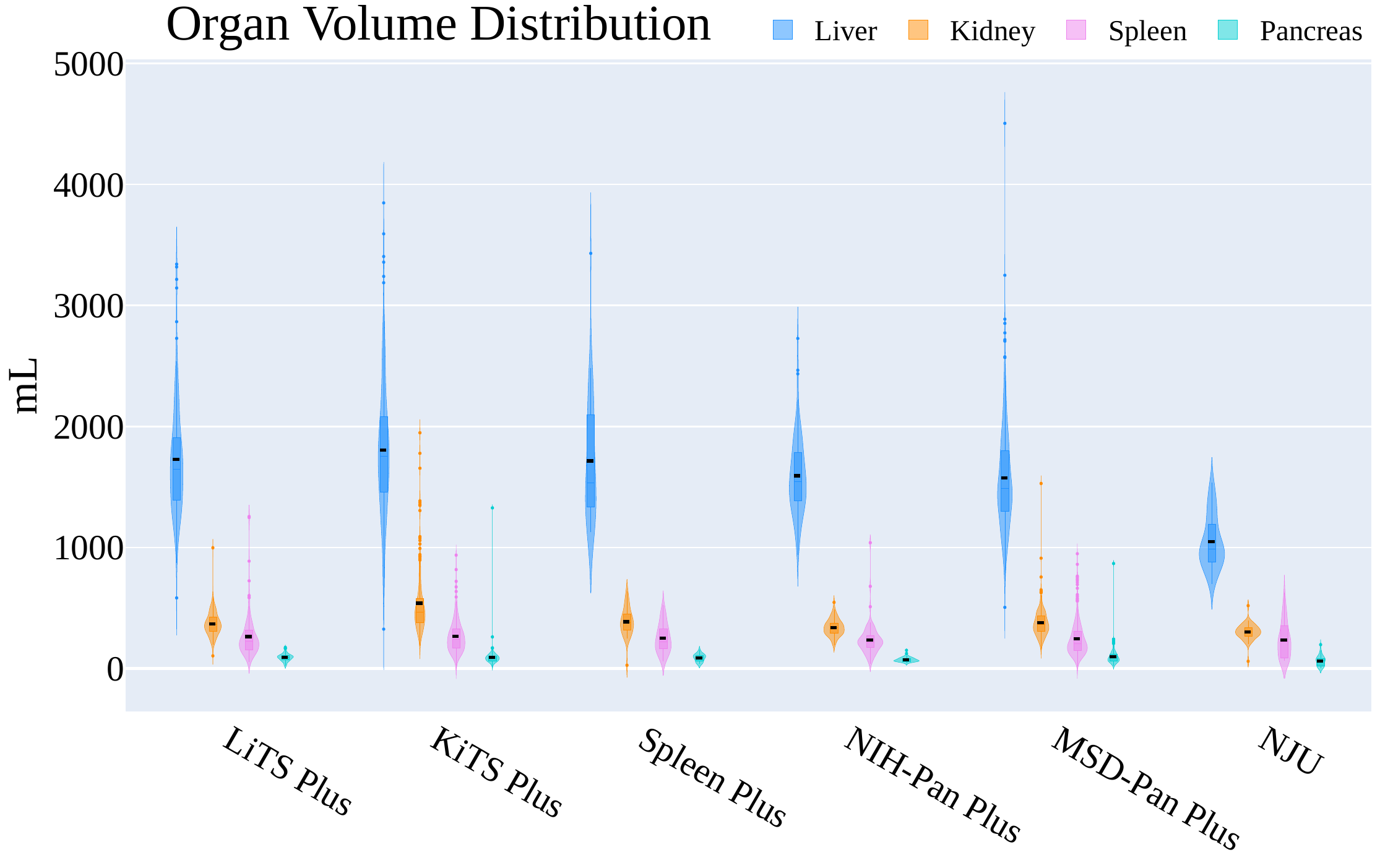}}\\
    	\subfloat{\includegraphics[scale=0.38]{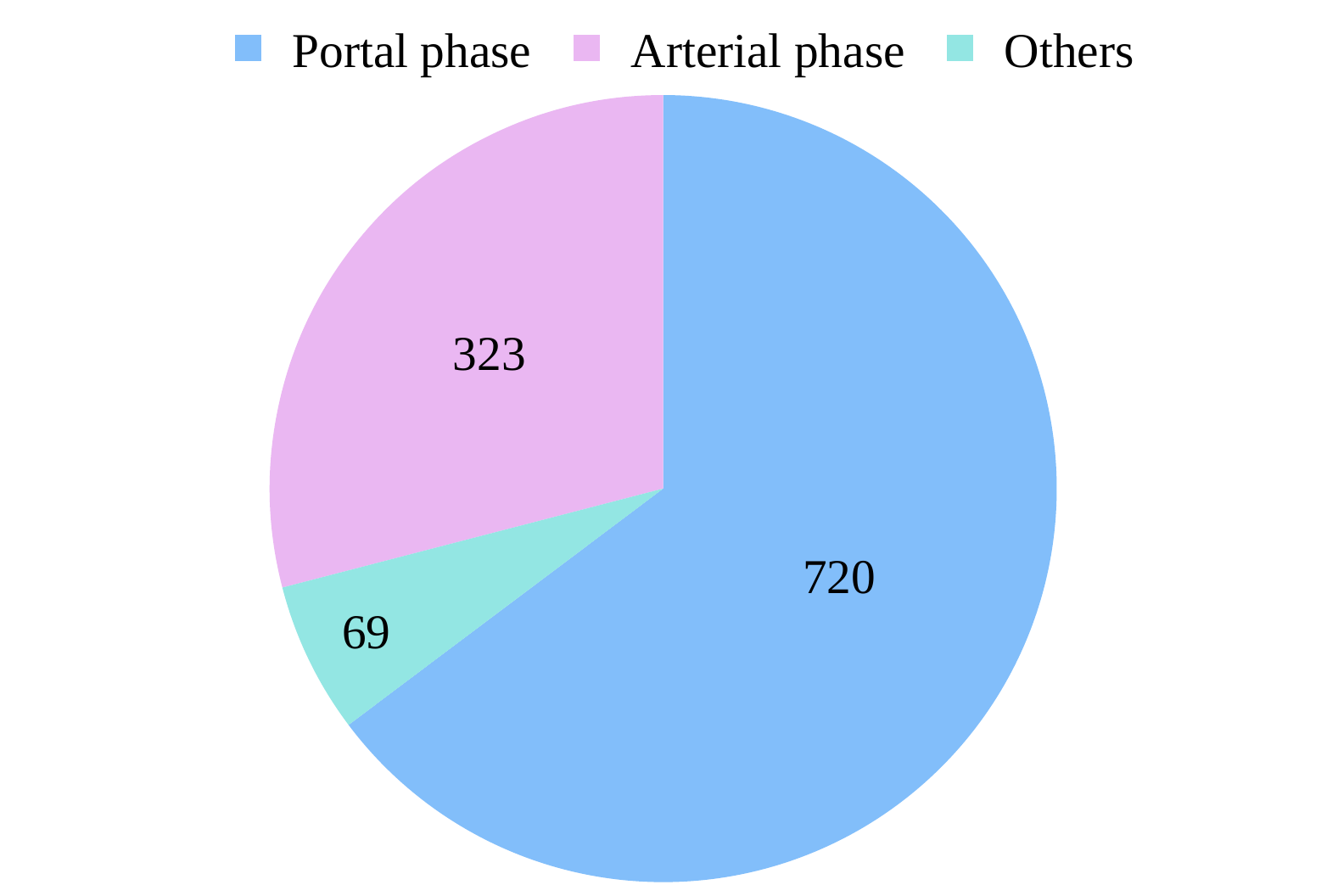}}\\
	\end{center}
	\caption{Organ volume and contrast phase distributions in AbdomenCT-1K.}\label{fig:distribution}
	\label{fig:}
\end{figure}

The original datasets only provide annotations of one single organ, while our dataset contains annotations of four organs for all cases in each dataset as shown in Figure~\ref{fig:dataoverview}.
In order to distinguish from the original datasets, we term our multi-organ annotations as plus datasets (e.g., the multi-organ LiTS dataset is termed as LiTS Plus dataset in this paper). Figure~\ref{fig:distribution} presents the organ volume and contrast phase distributions in AbdomenCT-1K. The other information (e.g., CT scanners, the distribution of the Hounsfield unit (HU) value, image size, and image spacing.) is presented in the supplementary (Supplementary Table 1).

\subsection{Annotation}
Annotations from the existing datasets are used if available, and we further annotate the absent organs in these datasets. Specifically, we first use the trained single-organ models to infer each case. Then, 15 junior annotators (one to five years of experience) use ITK-SNAP 3.6 to manually refine the segmentation results under the supervision of two board-certified radiologists. Finally, one senior radiologist with more than 10-years experience verifies and refines the annotations. All the annotations are applied to axial images. To reduce inter-rater annotation variability, we introduce three hierarchical strategies to improve the label consistency. Specifically, 
\begin{itemize}
    \item before annotation, all raters are required to learn the existing organ annotation protocols, aiming to ensure that the annotation protocols are consistent in raters and the existing datasets;
    \item during annotation, the obvious label errors in existing datasets are fixed and all annotations are finally checked and revised by an experienced senior radiologist (10+ years specialized in the abdomen);
    \item after annotation, we train five-fold cross-validation U-Net models to find the possible segmentation errors. The cases with low DSC or NSD scores are double-checked by the senior radiologist.
\end{itemize}

In addition, we invite two radiologists to annotate the 50 cases in the Nanjing University dataset and present their inter-rater variability in Table~\ref{tab:inter-rater}.

\begin{table}[!hp]
\caption{Quantitative analysis of inter-rater variability between two radiologists.}\label{tab:inter-rater}
\centering
\begin{tabular}{ccccc}
\hline
Organ    & Liver     & Kidney    & Spleen    & Pancreas  \\ \hline
DSC (\%) & 98.4 $\pm$ 0.52 & 98.7 $\pm$ 0.53 & 98.6 $\pm$ 0.84 & 93.8 $\pm$ 7.78 \\
NSD (\%) & 95.7 $\pm$ 3.04 & 98.7 $\pm$ 2.05 & 98.2 $\pm$ 4.18 & 92.5 $\pm$ 9.40 \\ \hline
\end{tabular}
\end{table}

\subsection{Backbone network}
The legendary U-Net (\cite{ronneberger20152DUNet, ronneberger20163DUNet}) has been widely used in various medical image segmentation tasks, and many variants have been proposed to improve it. However, recent studies \cite{isensee2020nnunet,KiTS} demonstrate that it is still hard to surpass a basic U-Net if the corresponding pipeline is designed adequately. In particular, nnU-Net (no-new-U-Net)  \cite{isensee2020nnunet} has been proposed to automatically adapt preprocessing strategies and network architectures (i.e., the number of pooling, convolutional kernel size, and stride size) to a given 3D medical dataset. Without manually tuning, nnU-Net can achieve better performances than most specialized deep learning pipelines in 19 public international segmentation competitions and set a new SOTA in 49 tasks.
Currently, nnU-Net is still the SOTA method in many segmentation tasks~\cite{ma2021SOTA-Seg}. Thus, we employ nnU-Net as our backbone network\footnote{The source code is publicly available at \url{https://github.com/MIC-DKFZ/nnUNet}.}.
Specifically, the network input is configured with a batch size of 2. The optimizer is stochastic gradient descent with an initial learning rate (0.01) and a nesterov momentum (0.99). To avoid overfitting, standard data augmentation techniques are used during training, such as rotation, scaling, adding Gaussian Noise, gamma correction. The loss function is a combination of Dice loss \cite{milletari2016Dice} and cross-entropy loss because compound loss functions have been proved to be robust in many segmentation tasks~\cite{SegLossOdyssey}.
All the models are trained for 1000 epochs with the above hyper-parameters on NVIDIA TITAN V100 or 2080Ti GPUs.

\subsection{Evaluation metrics}
Motivated by the evaluation methods of the well-known medical image segmentation decathlon\footnote{http://medicaldecathlon.com/}, we employ two complementary metrics to evaluate the segmentation performance. Specifically, Dice similarity coefficient (DSC), a region-based measure, is used to evaluate the region overlap. Normalized surface Dice (NSD) \cite{nikolov2018SDice}, a boundary-based measure, is used to evaluate how close the segmentation and ground truth surfaces are to each other at a specified tolerance $\tau$.
Both metrics take the scores in $[0,1]$ and higher scores indicate better segmentation performance.
Let $G, S$ denote the ground truth and the segmentation result, respectively. $|\partial G|$ and $|\partial S|$ are the number of voxels of the ground truth and the segmentation results, respectively.
We formulate the definitions of the two measures as follows:
\begin{itemize}
    \item {Region-based measure: DSC}
\begin{equation*}
    DSC(G, S) = \frac{2|G\cap S|}{|G| + |S|},
\end{equation*}
\item {Boundary-based measure: NSD}
\begin{equation*}
    NSD(G, S) = \frac{|\partial G\cap B_{\partial S}^{(\tau)}| + |\partial S\cap B_{\partial G}^{(\tau)}|}{|\partial G| + |\partial S|},
\end{equation*}
\end{itemize}
where  $B_{\partial G}^{(\tau)} = \{x\in R^3 \, | \, \exists \tilde{x}\in \partial G,\, ||x-\tilde{x}||\leq \tau \}$, $B_{\partial S}^{(\tau)} = \{x\in R^3 \,|\, \exists \tilde{x}\in \partial S,\, ||x-\tilde{x}||\leq \tau \}$  denote the border region of the ground truth and the segmentation surface at tolerance $\tau$, respectively. In this paper, we set the tolerance $\tau$ as $1$mm.

\begin{figure}[!htbp]
\begin{center}
    \includegraphics[scale=0.38]{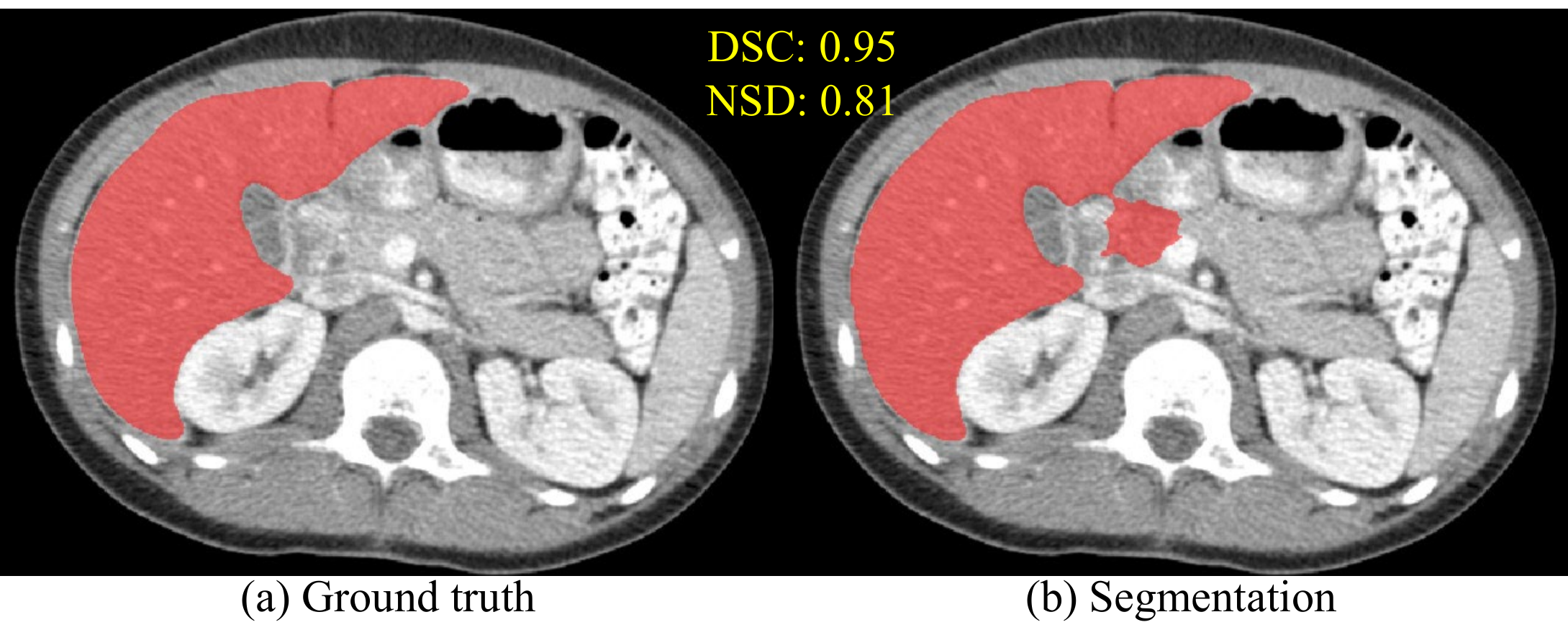}
\end{center}
\caption{Comparison of Dice similarity coefficient (DSC) and normalized surface Dice (NSD).}
\label{fig:NSD-vs-DSC}
\end{figure}

DSC is a commonly used segmentation metric and has been used in many segmentation benchmarks~\cite{bilic2019lits, KiTS}, while NSD can provide additional complementary information to the segmentation quality.
Figure~\ref{fig:NSD-vs-DSC} presents a liver segmentation example to illustrate the features of NSD. An obvious segmentation error can be found on the right side boundary of the liver. However, the DSC score is still very high that cannot well reflect the boundary error, while NSD is sensitive to this boundary error and thus a low score is obtained. In many clinical tasks, such as preoperative planning and organ transplant, boundary errors are critical~\cite{meinzer2002LiverClinical, ni2020LiverResection} and thus should be eliminated.
Another benefit of introducing NSD is that it ignores small boundary deviations because small inter-observer errors are also unavoidable and often not clinically relevant when segmenting the organs by radiologists.
In all the experiments, we employ the official implementation at \url{http://medicaldecathlon.com/files/Surface_distance_based_measures.ipynb} to compute the metrics.

\section{A large-scale study on fully supervised organ segmentation}
\label{S:large-scale}
Abdominal organ segmentation is one of the most popular segmentation tasks. Most of the existing benchmarks mainly focus on fully supervised segmentation tasks and are built on single-center datasets where training cases and testing cases are from the same medical centers, and the state-of-the-art (SOTA) method (nnU-Net \cite{isensee2020nnunet}) has achieved very high accuracy.
In this section, we evaluate the SOTA method on our plus datasets to show whether the performance can generalize to multi-center datasets.

\subsection{Single organ segmentation}
Existing abdominal organ segmentation benchmarks mainly focus on single organ segmentation, such as KiTS, MSD-Spleen, and NIH Pancreas only focus on kidney segmentation, spleen segmentation, and pancreas segmentation, respectively.
The training and testing sets in these benchmarks are from the same medical center, and the current SOTA method has achieved human-level accuracy (in terms of DSC) in some tasks (i.e., liver segmentation, kidney segmentation, and spleen segmentation). However, it is unclear whether the great performance can generalize to new datasets from third-party medical centers.
In this subsection, we randomly select 80\% of cases for training in the original training set and the remaining 20\% of cases and three new datasets as testing set, which can allow quantitative comparisons within-dataset and across-dataset.

\begin{table}[!htp]
\centering
\caption{Quantitative results of single organ segmentation. Each segmentation task has one testing set from the same data source as the training set and three testing sets from new medical centers. The bold and underlined numbers denote the best and worst results, respectively.}
\label{tab:exp-single}
\renewcommand\tabcolsep{2.5pt}
\begin{tabular}{lllcc} 
\hline
Task     & Training       & Testing                                                                                          & \multicolumn{1}{c}{DSC (\%)}                                                                            & \multicolumn{1}{c}{NSD (\%)}                                                                             \\ 
\hline
Liver    & LiTS (104)     & \begin{tabular}[c]{@{}l@{}}LiTS (27)\\ KiTS (210) \\Spleen (41) \\Pancreas (361)\end{tabular}     & \begin{tabular}[c]{@{}l@{}}\textbf{97.4$\pm$0.63}\\\underline{94.9$\pm$7.59}\\96.5$\pm$3.31\\96.4$\pm$3.07\end{tabular} & \begin{tabular}[c]{@{}l@{}}\underline{83.2$\pm$5.89}\\\underline{83.2$\pm$12.2}\\\textbf{86.6$\pm$7.54}\\85.4$\pm$8.46\end{tabular}  \\ 
\hline
Kidney   & KiTS (168)     & \begin{tabular}[c]{@{}l@{}}KiTS (42) \\LiTS (131)\\Pancreas (361)\\Spleen (41)\end{tabular}      & \begin{tabular}[c]{@{}l@{}}\textbf{97.1$\pm$3.81}\\87.5$\pm$17.9\\\underline{82.0$\pm$28.9}\\93.7$\pm$6.52\end{tabular} & \begin{tabular}[c]{@{}l@{}}\textbf{94.0$\pm$6.91}\\\underline{75.0$\pm$16.5}\\\underline{75.0$\pm$27.1}\\82.5$\pm$9.97\end{tabular}  \\ 
\hline
Spleen   & Spleen (33)    & \begin{tabular}[c]{@{}l@{}}Spleen Ts (8) \\LiTS (131)\\KiTS (210)\\Pancreas (361)\\\end{tabular} & \begin{tabular}[c]{@{}l@{}}\textbf{97.2$\pm$0.81}\\91.0$\pm$15.5\\\underline{86.6$\pm$23.3}\\94.6$\pm$8.32\end{tabular} & \begin{tabular}[c]{@{}l@{}}\textbf{94.6$\pm$4.41}\\79.6$\pm$16.4\\\underline{76.7$\pm$23.7}\\86.9$\pm$10.4\end{tabular}  \\ 
\hline
Pancreas & MSD Pan. (225) & \begin{tabular}[c]{@{}l@{}}MSD Pan. (56)\\LiTS (131)\\KiTS (210)\\Spleen (41)\end{tabular}       & \begin{tabular}[c]{@{}l@{}}86.1$\pm$6.59\\86.6$\pm$12.2\\\underline{80.9$\pm$10.5}\\\textbf{86.6$\pm$8.80}\end{tabular} & \begin{tabular}[c]{@{}l@{}}66.1$\pm$15.4\\75.4$\pm$14.2\\\underline{61.5$\pm$12.2}\\\textbf{77.7$\pm$11.6}\end{tabular}  \\ 
\hline
\end{tabular}
\end{table}

\begin{figure}[htbp]
	\begin{center}
    	\subfloat{\includegraphics[scale=0.25]{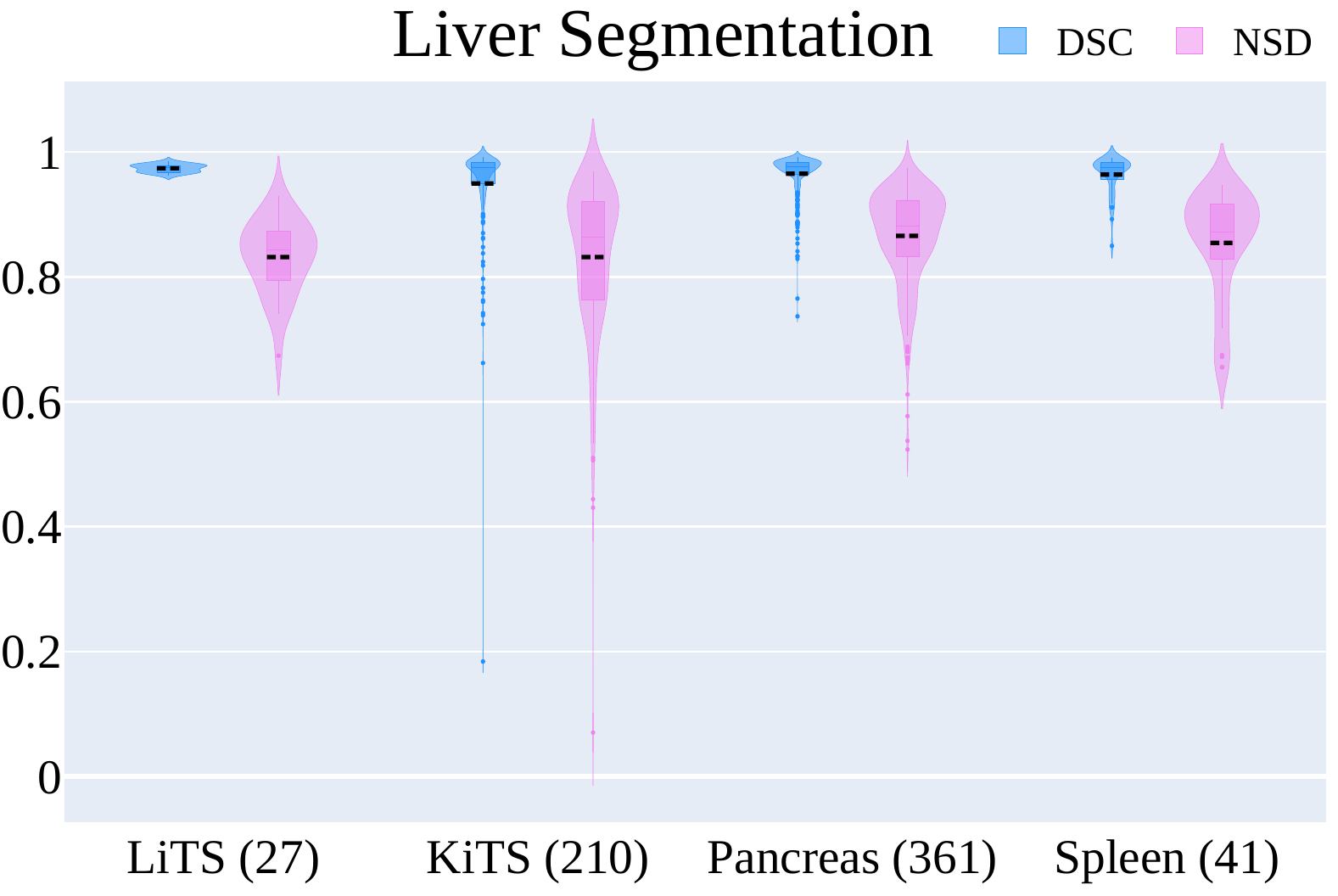}}
    	\subfloat{\includegraphics[scale=0.25]{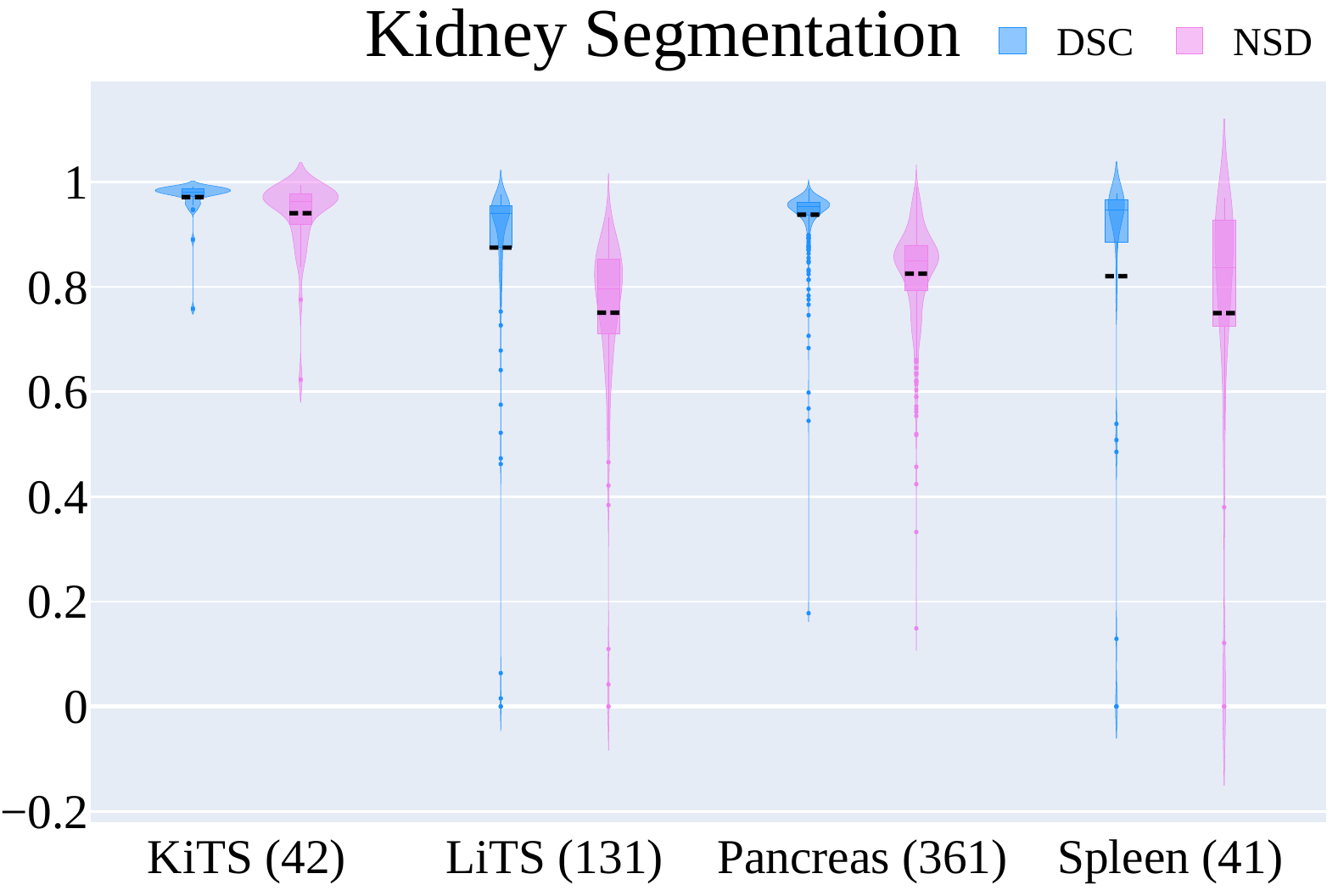}}\\
     	\subfloat{\includegraphics[scale=0.25]{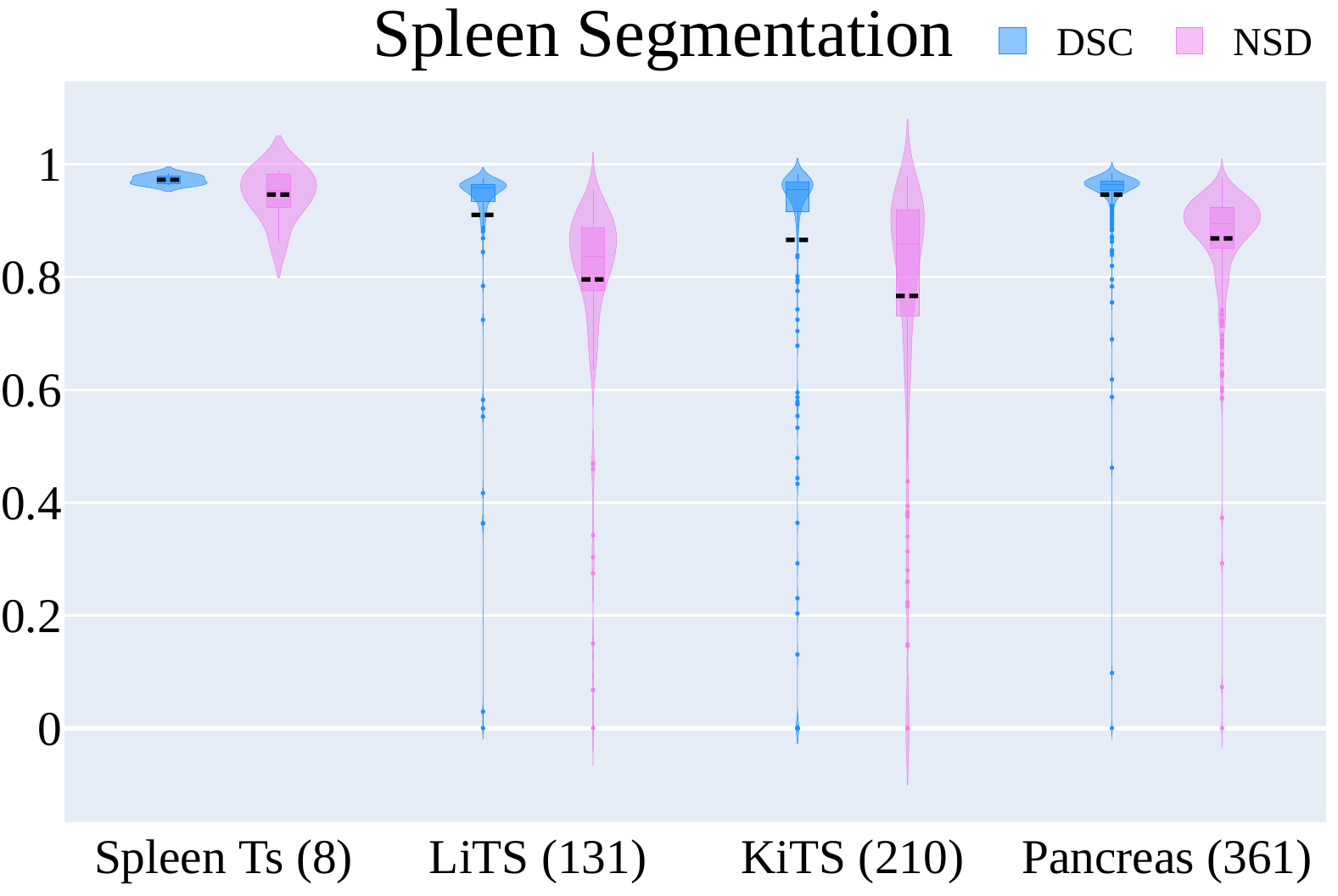}}
    	\subfloat{\includegraphics[scale=0.25]{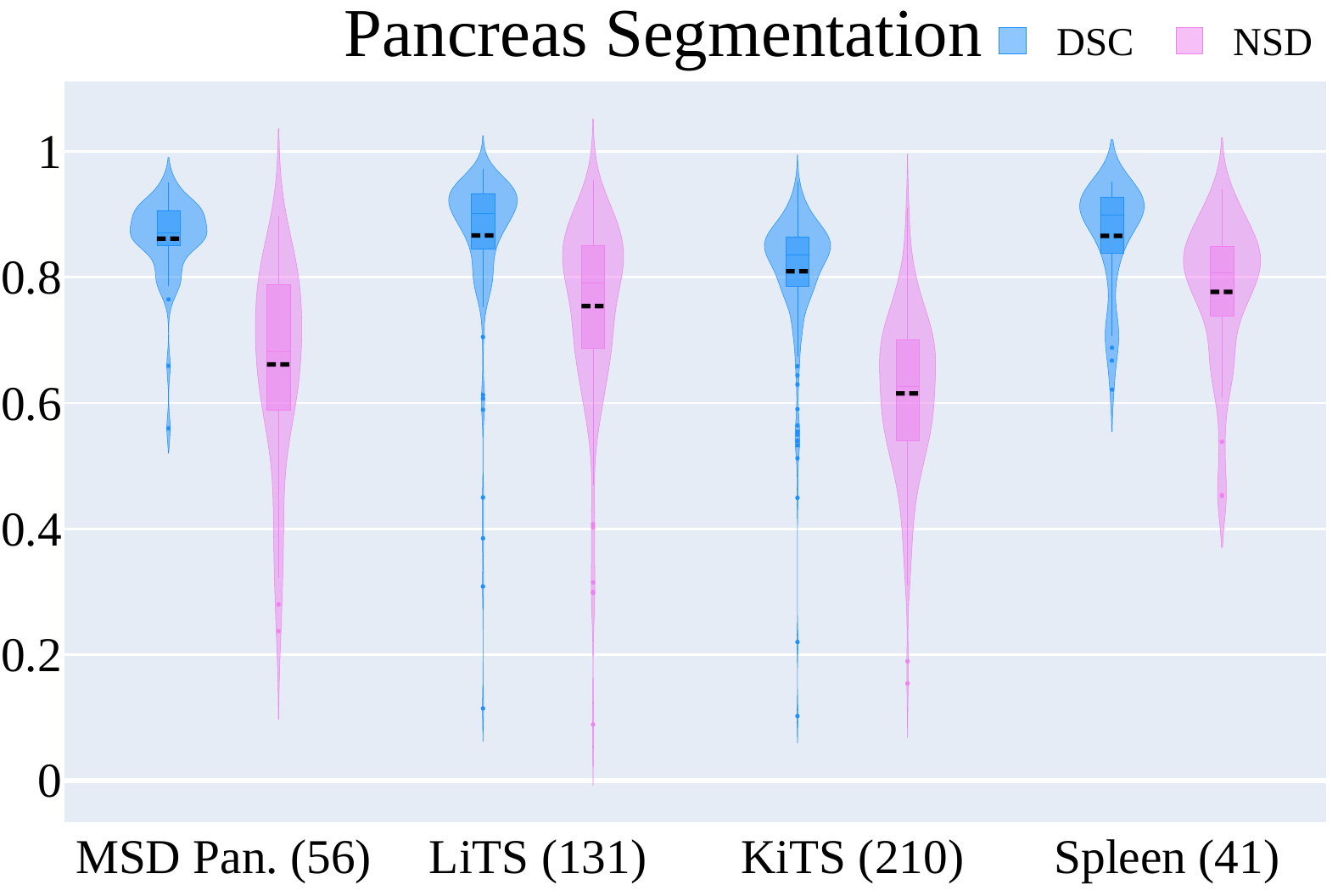}}\\
	\end{center}
	\caption{Violin plots of the segmentation performances (DSC and NSD) of different organs in single organ segmentation tasks.}\label{fig:sec3-SingleOrgan}
\end{figure}

\begin{table*}[!ht]
\centering
\caption{Quantitative results of fully supervised multi-organ segmentation in terms of average DSC and NSD. The bold and underlined numbers denote the best and worst results, respectively.}
\label{tab:exp-4organ}
\renewcommand\tabcolsep{3pt}
\begin{tabular}{llcccccccc}
\hline
\multirow{2}{*}{Training} & \multirow{2}{*}{Testing}                                                                           & \multicolumn{2}{c}{Liver}                                         & \multicolumn{2}{c}{Kidney}                                        & \multicolumn{2}{c}{Spleen}                                        & \multicolumn{2}{c}{Pancreas}                                       \\
\cline{3-10}
                          &                                                                                                    & DSC (\%)                                                         & NSD (\%) & DSC (\%)                                                         & NSD (\%) & DSC (\%)                                                         & NSD (\%) & DSC (\%)                                                         & NSD (\%)  \\
\hline
LiTS Plus (131)      & \begin{tabular}[c]{@{}l@{}}KiTS Plus (210) \\Spleen Plus (41)\\Pancreas Plus (361)\end{tabular}    & \begin{tabular}[c]{@{}c@{}}97.1$\pm$3.42\\96.9$\pm$4.66\\\textbf{98.2$\pm$1.39}\end{tabular} & \begin{tabular}[c]{@{}c@{}}88.6$\pm$10.3\\89.1$\pm$11.2\\91.9$\pm$5.77\end{tabular}    & \begin{tabular}[c]{@{}c@{}}89.1$\pm$14.5\\85.6$\pm$26.7\\\textbf{96.0$\pm$5.04}\end{tabular} & \begin{tabular}[c]{@{}c@{}}81.9$\pm$13.5\\\underline{78.9$\pm$26.0}\\\textbf{92.4$\pm$7.06}\end{tabular}    & \begin{tabular}[c]{@{}c@{}}\underline{92.6$\pm$13.6}\\95.0$\pm$11.5\\97.5$\pm$5.88\end{tabular} &  \begin{tabular}[c]{@{}c@{}}\underline{86.0$\pm$16.3}\\91.6$\pm$12.3\\96.0$\pm$7.22\end{tabular}   & \begin{tabular}[c]{@{}c@{}}84.7$\pm$8.63\\86.1$\pm$15.6\\81.1$\pm$10.7\end{tabular} &   \begin{tabular}[c]{@{}c@{}}70.4$\pm$10.7\\78.8$\pm$16.2\\61.4$\pm$13.3\end{tabular}   \\
\hline
KiTS Plus (210)      & \begin{tabular}[c]{@{}l@{}}LiTS Plus (131)\\Spleen Plus (41)\\Pancreas Plus (361)\end{tabular}     & \begin{tabular}[c]{@{}c@{}}\underline{95.5$\pm$3.93}\\97.1$\pm$4.26\\98.0$\pm$2.67\end{tabular}     &   \begin{tabular}[c]{@{}c@{}}\underline{77.4$\pm$8.97}\\90.4$\pm$6.20\\91.3$\pm$6.55\end{tabular}  & \begin{tabular}[c]{@{}c@{}}91.4$\pm$13.2\\\underline{84.9$\pm$25.4}\\94.4$\pm$5.61\end{tabular}     &  \begin{tabular}[c]{@{}c@{}}79.3$\pm$14.0\\79.2$\pm$23.6\\84.3$\pm$9.22\end{tabular}   & \begin{tabular}[c]{@{}c@{}}95.0$\pm$10.6\\96.6$\pm$1.92\\96.8$\pm$6.24\end{tabular}     &   \begin{tabular}[c]{@{}c@{}}91.6$\pm$11.3\\93.8$\pm$4.28\\94.9$\pm$7.96\end{tabular}  & \begin{tabular}[c]{@{}c@{}}87.4$\pm$10.9\\85.6$\pm$14.8\\80.5$\pm$11.5\end{tabular}     &   \begin{tabular}[c]{@{}c@{}}74.9$\pm$12.9\\76.7$\pm$15.5\\61.5$\pm$16.9\end{tabular}   \\
\hline
MSD Pan. Plus (281)  & \begin{tabular}[c]{@{}l@{}}LiTS Plus (131)\\KiTS Plus (210)\\Spleen Plus (41)\end{tabular} & \begin{tabular}[c]{@{}c@{}}96.2$\pm$2.58\\98.0$\pm$3.19\\98.1$\pm$1.68\end{tabular} &  \begin{tabular}[c]{@{}c@{}}77.8$\pm$7.09\\\textbf{92.1$\pm$10.3}\\91.4$\pm$6.04\end{tabular}   & \begin{tabular}[c]{@{}c@{}}94.7$\pm$9.22\\90.8$\pm$11.2\\94.8$\pm$3.71\end{tabular} &  \begin{tabular}[c]{@{}c@{}}89.7$\pm$11.7\\82.7$\pm$12.4\\87.8$\pm$5.47\end{tabular}   & \begin{tabular}[c]{@{}c@{}}96.3$\pm$9.19\\94.6$\pm$12.3\\\textbf{98.5$\pm$0.86}\end{tabular} &  \begin{tabular}[c]{@{}c@{}}93.0$\pm$10.5\\88.5$\pm$15.3\\\textbf{97.0$\pm$3.38}\end{tabular}   & \begin{tabular}[c]{@{}c@{}}\textbf{90.1$\pm$10.5}\\\underline{80.0$\pm$14.1}\\88.2$\pm$8.44\end{tabular} &   \begin{tabular}[c]{@{}c@{}}\textbf{82.3$\pm$13.2}\\63.1$\pm$12.9\\80.9$\pm$12.7\end{tabular}   \\
\hline
\multirow{3}{*}{Spleen Plus (41)} & LiTS Plus (131)                                                                                 & \multicolumn{1}{l}{96.0$\pm$3.35}                                                                & \multicolumn{1}{l}{78.2$\pm$7.50}                                                                & \multicolumn{1}{l}{95.2$\pm$5.88}                                                                & \multicolumn{1}{l}{85.9$\pm$7.19}                                                                & \multicolumn{1}{l}{95.3$\pm$9.85}                                                                & \multicolumn{1}{l}{90.6$\pm$11.6}                                                                & \multicolumn{1}{l}{88.8$\pm$9.61}                                                                & \multicolumn{1}{l}{79.9$\pm$11.8}                                                                 \\
                                  & Pancreas Plus (361)                                                                              & \multicolumn{1}{l}{97.9$\pm$2.43}                                                                & \multicolumn{1}{l}{91.2$\pm$6.21}                                                                & \multicolumn{1}{l}{95.4$\pm$5.45}                                                                & \multicolumn{1}{l}{87.5$\pm$6.69}                                                                & \multicolumn{1}{l}{97.7$\pm$3.63}                                                                & \multicolumn{1}{l}{96.1$\pm$5.75}                                                                & \multicolumn{1}{l}{80.2$\pm$12.5}                                                                & \multicolumn{1}{l}{\underline{60.7$\pm$14.0}}                                                                 \\
                                  & KiTS Plus (210)                                                                                 & \multicolumn{1}{l}{96.8$\pm$4.80}                                                                & \multicolumn{1}{l}{89.7$\pm$9.77}                                                                & \multicolumn{1}{l}{89.7$\pm$16.5}                                                                & \multicolumn{1}{l}{85.0$\pm$15.0}                                                                & \multicolumn{1}{l}{93.7$\pm$13.1}                                                                & \multicolumn{1}{l}{86.4$\pm$15.9}                                                                & \multicolumn{1}{l}{83.3$\pm$10.7}                                                                & \multicolumn{1}{l}{68.9$\pm$12.1}                                                                 \\                                  
\hline
\end{tabular}
\end{table*}

Table~\ref{tab:exp-single} shows the quantitative segmentation results for each organ and Figure~\ref{fig:sec3-SingleOrgan} shows the corresponding violin plots. It can be found that
\begin{itemize}
  \item for liver segmentation, the SOTA method achieves high DSC scores ranging from 94.9\% to 96.5\% on the three new testing datasets, demonstrating its good generalization ability. Compared to the DSC scores on LiTS (27) testing set, the DSC scores drop 2.5\% on the KiTS (210). The main reason is that the CT scans in KiTS (210) were acquired on the arterial phase while most of the CT scans in LiTS were acquired on the portal phase. Both Pancreas (361) and Spleen (41) obtain relatively close DSC scores compared with the LiTS (27), but the NSD scores are much better, indicating that the segmentation results in LiTS (27) have more errors near the boundary. This is because most cases in LiTS have liver cancers while most cases in Pancreas (361) and Spleen (41) are normal in liver.
  \item for kidney segmentation, compared with the high DSC and NSD scores on KiTS (42), the performance drops remarkably on the other three datasets with up to 15\% in DSC and 19\% in NSD, especially for the LiTS (131) and the Pancreas (361). The main reason is that the CT phases of most cases in the other three datasets are different from the KiTS.
  \item for spleen segmentation, both DSC and NSD scores also drop on the other three datasets, especially for the KiTS (210) datasets where 10.6\% dropping in DSC and 17.9\%  dropping in NSD is observed, indicating that the SOTA method does not generalize well on different CT phases.
  \item for pancreas segmentation, the performance also has a significant decline on KiTS (210) because of the differences in CT phases. Remarkably, the LiTS (131) and Spleen (41) obtain similar DSC scores compared to the MSD Pan. (56), but the NSD scores have large improvements with 9.3\% and 11\% because most cases in the two datasets have a healthy pancreas. The results demonstrate that the pancreas segmentation model generalizes better on pancreas healthy cases than pancreas pathology cases, especially for the boundary-based metric NSD.
\end{itemize}
In summary, the current SOTA single organ segmentation method can achieve very high performance (especially for the DSC) when the training set and the testing set are from the same distribution, but the high performance would degrade when the testing sets are from new medical centers.

\begin{figure*}[!ht]
\centering
\includegraphics[scale=0.5]{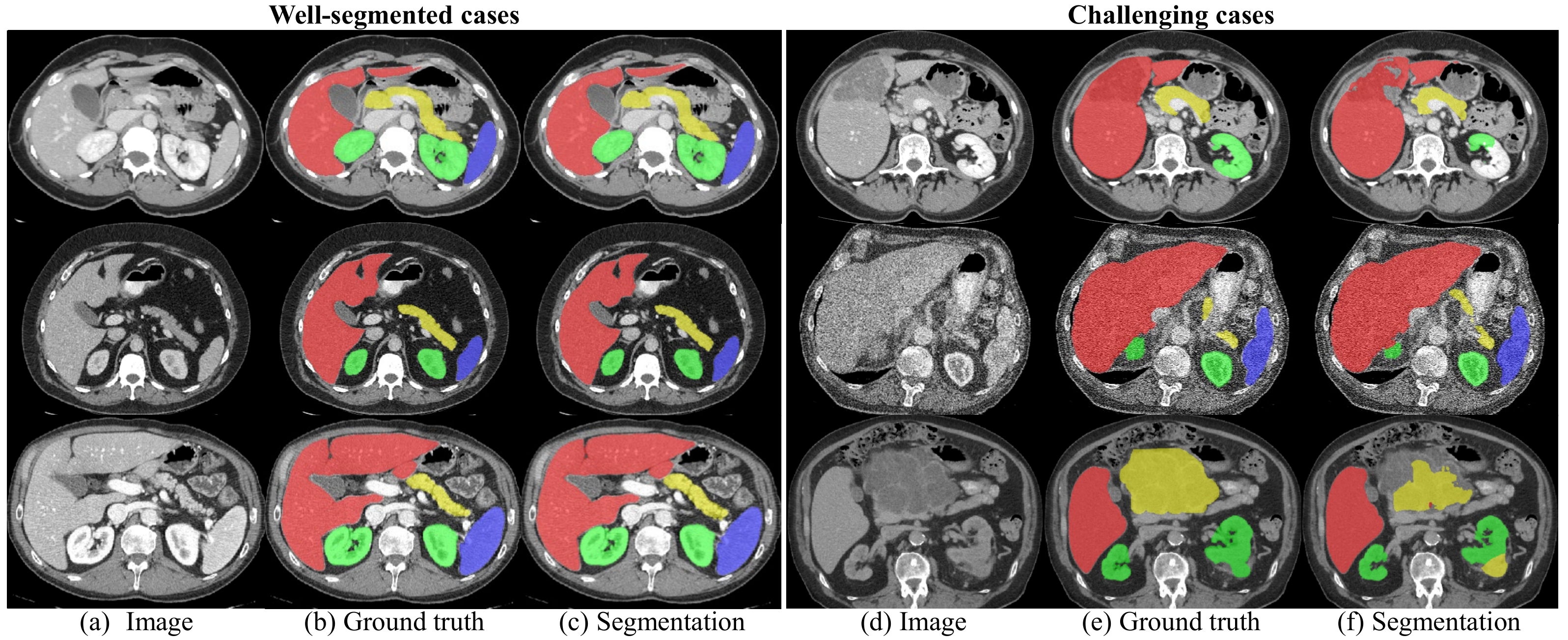}
\caption{Well-segmented and challenging examples from testing sets in the large-scale fully supervised multi-organ segmentation study.}
\label{fig:3-2-examples}
\end{figure*}

\subsection{Multi-organ segmentation}
In this subsection, we focus on evaluating the generalization ability of the SOTA method (nnU-Net) on multi-organ segmentation tasks. Specifically, we conduct four groups of experiments. In each group, we train the nnU-Net on one dataset with four organ annotations and test the trained model on the other three new datasets. It should be noted that the training set and testing set are from different medical centers in each group.

Table~\ref{tab:exp-4organ} shows quantitative segmentation results for each organ\footnote{The corresponding violin plots are presented in supplementary (Supplementary Figure 1).}. It can be observed that
\begin{itemize}
  \item the DSC scores are relatively stable in liver and spleen segmentation results, achieving $90\%+$ in all experiments. However, the NSD scores fluctuate greatly among different testing sets, ranging from 77.4\% to 92.1\% for liver segmentation and from 86.0\% to 97.0\% for spleen segmentation.
  \item both DSC and NSD scores vary greatly in kidney segmentation results for different testing sets. For example, in the first group experiments, nnU-Net achieves average kidney DSC scores of 96.0\% and 85.6\%, and NSD scores of 92.4\% and 78.9\% on Pancreas Plus (361) and Spleen Plus (41) datasets, respectively, which has a  performance gap of 10\%+.
  \item pancreas segmentation results are lower than the other organs across all experiments, indicating that pancreas segmentation is still a challenging problem.
\end{itemize}

Figure~\ref{fig:3-2-examples} presents some examples of well-segmented and challenging cases. It can be observed that the well-segmented cases have clear boundaries and good contrast for the organs, and there exist no severe artifacts or lesions in the organs. In contrast with the well-segmented cases, the challenging cases usually have heterogeneous lesions, such as the liver lesion (Figure~\ref{fig:3-2-examples} (d)-1$^{st}$ row) and the pancreas lesions (Figure~\ref{fig:3-2-examples} (d)-3$^{rd}$ row). In addition, the image quality can be degraded by the noise, e.g., Figure~\ref{fig:3-2-examples} (d)-2$^{nd}$ row.

\subsection{Is abdominal organ segmentation a solved problem?}
In summary, for the question:
\begin{center}
    \emph{Is abdominal organ segmentation a solved problem?}
\end{center}
 the answer would be \textbf{Yes} for liver, kidney, and spleen segmentation, if
\begin{itemize}
    \item the evaluation metric is DSC, which mainly focuses on evaluating the region-based segmentation error.
    \item the data distribution of the testing set is the same as the training set.
    \item the cases in the testing set are trivial, which means that the cases do not have severe diseases and low image quality.
\end{itemize}

However, we argue that \textbf{abdominal organ segmentation  remains to be an unsolved problem} in following situations:
\begin{itemize}
    \item the evaluation metric is NSD, which focuses on evaluating the accuracy of organ boundaries.
    \item testing sets are from new medical centers with different data distributions from the training set.
    \item the cases in the testing sets have unseen or severe diseases and low image quality, such as heterogeneous lesions and noise, while training sets do not have or only have few similar cases.
\end{itemize}

As mentioned in Section~\ref{SS:existing_dataset}, existing abdominal organ segmentation benchmarks cannot reflect these challenging situations. Thus, in this work, we build new segmentation benchmarks that can cover these challenges.
Existing benchmarks have received extensive attention in the community and have little rooms for improvements in current testing sets and associated evaluation metric (i.e., DSC)~\cite{bilic2019lits,KiTS}. Therefore, we expect that our new segmentation benchmarks would bring new insights and again attract wide attention.


\section{New abdominal CT organ segmentation benchmarks on Fully supervised, semi-supervised, weakly supervised and continual learning}
\label{S:benchmark}
Our new abdominal organ segmentation benchmarks aim to include more challenging settings. In particular, we focus on
\begin{itemize}
    \item evaluating not only region related segmentation errors but also boundary related segmentation errors, because the boundary errors are critical in many clinical applications, such as surgical planning for organ transplantation.
    \item evaluating the generalization ability of segmentation methods on cases from new medical centers and CT phases.
    \item evaluating the generalization ability of segmentation methods on cases with unseen and severe diseases.
\end{itemize}

In addition to the fully supervised segmentation benchmark, we also set up, to the best of our knowledge, the first abdominal organ segmentation benchmarks for semi-supervised learning, weakly supervised learning, and continual learning, which are currently active research topics and can alleviate the dependency on annotations. In each benchmark, we select 50 challenging cases and 50 random cases as the testing set, which is friendly to future users to evaluate their methods because it does not cost too much time during inference. More importantly, the final performance is not easy to be biased by the easy cases. We also introduce a new dataset as the common testing set, which can allow apple-to-apple comparisons among the four benchmarks.
Moreover, for each benchmark, we have developed a strong baseline with SOTA methods, which can be an out-of-the-box method for researchers who are interested in these tasks.

\subsection{Fully supervised abdominal organ segmentation benchmark}
Fully supervised segmentation is a long-term and popular research topic. In this benchmark, we focus on multi-organ segmentation (liver, kidney, spleen, and pancreas) and aim to deal with the unsolved problems that are presented in the large-scale study in Section \ref{S:large-scale}.

\begin{table*}
\centering
\caption{Task settings and quantitative baseline results of fully supervised multi-organ segmentation benchmark.}
\label{tab:fullSup-benchmark}
\resizebox{\textwidth}{!}{
\begin{tabular}{cccccccccc} 
\hline
\multirow{2}{*}{Training}                                                                                & \multirow{2}{*}{Testing}                                                                                           & \multicolumn{2}{c}{Liver}       & \multicolumn{2}{c}{Kidney}      & \multicolumn{2}{c}{Spleen}      & \multicolumn{2}{c}{Pancreas}     \\ 
\cline{3-10}
                                                                                                         &                                                                                                                    & DSC (\%)            & NSD (\%)            & DSC (\%)            & NSD (\%)            & DSC (\%)            & NSD (\%)            & DSC (\%)            & NSD (\%)             \\ 
\hline
\begin{tabular}[c]{@{}c@{}}MSD Pan. Plus (281) \\NIH Pan. Plus (80)\\Subtask 1: 361 cases\\\end{tabular}          & \multirow{2}{*}{\begin{tabular}[c]{@{}c@{}} 100 cases \end{tabular}} & 95.8$\pm$6.04  & 83.0$\pm$12.1  & 84.1$\pm$14.8  & 73.8$\pm$14.1  & 89.8$\pm$15.5  & 80.6$\pm$18.3  & 65.0$\pm$22.7  & 55.2$\pm$17.6   \\ 
\cline{1-1}\cline{3-10}
\begin{tabular}[c]{@{}c@{}}MSD Pan. Plus (281)\\LiTS Plus (40)\\KiTS Plus (40)\\Subtask 2: 361 cases\end{tabular} &                                                                                                                    & \textbf{97.0$\pm$2.93}  & \textbf{85.8$\pm$9.92}  & \textbf{91.7$\pm$11.6}  & \textbf{84.1$\pm$13.1}  & \textbf{93.6$\pm$13.3}  & \textbf{87.4$\pm$15.0}  & \textbf{78.1$\pm$15.8}  & \textbf{65.0$\pm$15.2}   \\
\hline
\end{tabular}}
\end{table*}

\subsubsection{Task setting}
\textbf{Motivation of the training set and the testing set choice:} a large training set is often expected in fully supervised organ segmentation. Thus, we choose MSD Pan. Plus (281) as the base dataset in the training set because it has the largest number of training cases. On top of MSD Pan. Plus (281), different cases are added to the training set to build two subtasks as shown in Table~\ref{tab:fullSup-benchmark}. 
\begin{itemize}
    \item \textbf{Subtask 1.} The training set is composed of MSD Pan. Plus (281) and NIH Pan. Plus (80) where all the CT scans are from the portal phase. We use the baseline model in Section \ref{ss:fullySup-baseline} to predict all the remaining cases in LiTS Plus, KiTS Plus, and Spleen Plus. Then, 50 cases with the lowest average DSC and NSD are selected as the testing set. These cases usually have heterogeneous lesions and unclear boundaries, which are very challenging to segment and also very important in clinical practice.
    \item \textbf{Subtask 2.} The added NIH Pan. Plus (80) is replaced by 40 cases from LiTS Plus and 40 cases from KiTS Plus that have similar phases as the testing set. In this way, one can evaluate whether including shared contrast phases across training and testing sets can improve the performance or not.
\end{itemize}
We use the baseline model in Section~\ref{ss:fullySup-baseline} to infer all the remaining cases and select 100 cases as the final testing set, including 50 challenging cases with the lowest average DSC and NSD scores and 50 randomly selected cases. 
More importantly, the cases in the training set and the testing set have no overlap in each subtask.


\subsubsection{Baseline and results}
\label{ss:fullySup-baseline}
The baseline is built on 3D nnU-Net \cite{isensee2020nnunet}, which is the SOTA method for multi-organ segmentation. Table~\ref{tab:fullSup-benchmark} presents the detailed results for each organ in each subtask. It can be found that the performances of all organs in subtask 1 are lower than the performances in subtask 2 because the cases with shared contrast phases are introduced in subtask 2. Although fully supervised abdominal organ segmentation seems to be a solved problem (e.g., liver, kidney, and spleen segmentation) because SOTA methods have achieved inter-expert accuracy~\cite{kavur2020chaos, KiTS}, our studies on a large and diverse dataset demonstrate that abdominal organ segmentation is still an unsolved problem, especially for the challenging cases and situations.

\begin{figure}[!htbp]
\centering
\includegraphics[scale=0.4]{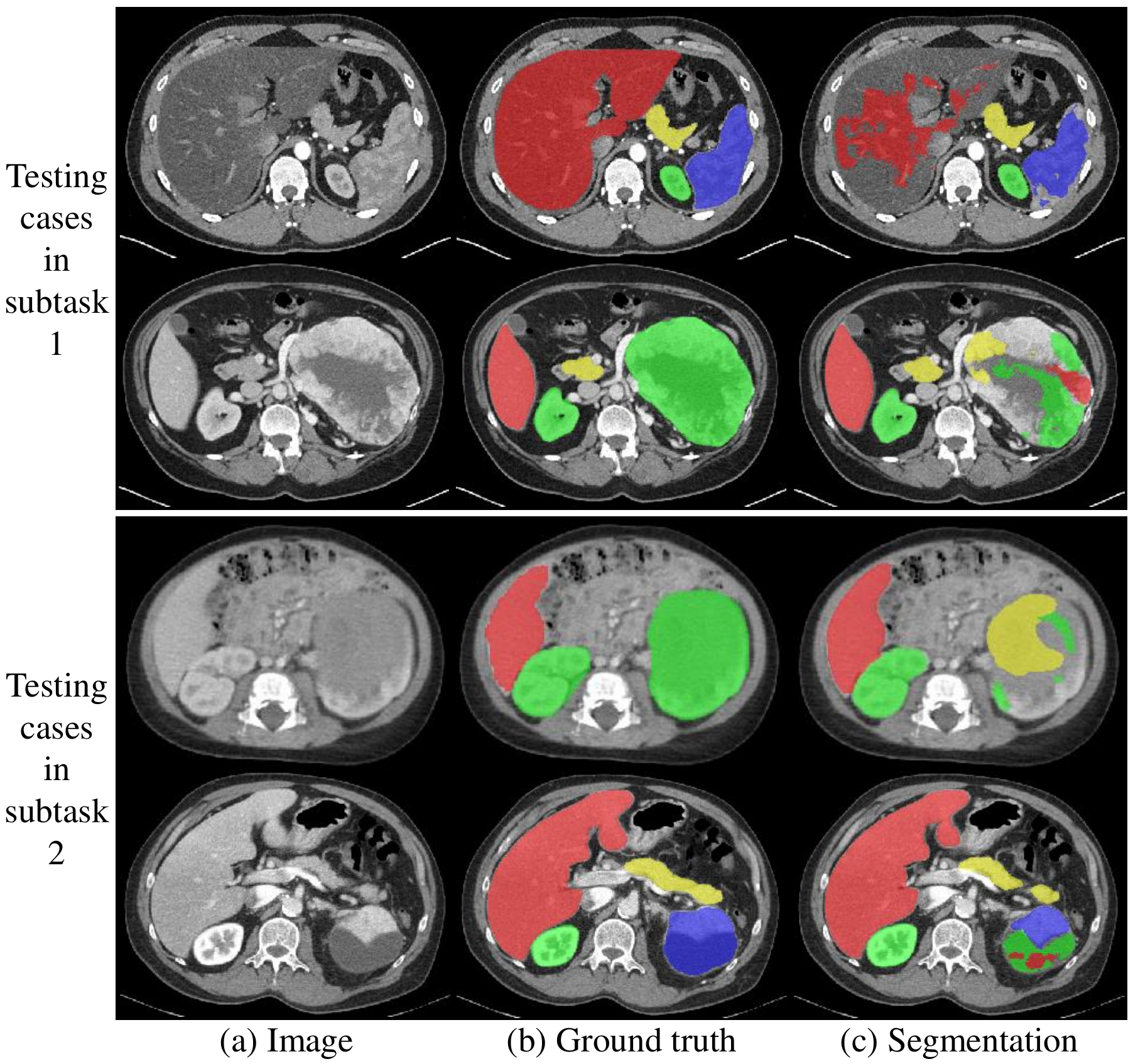}
\caption{Challenging examples from testing sets in fully supervised segmentation benchmark.}
\label{fig:4-1-exp}
\end{figure}

The violin plots of each organ are presented in Supplementary Figure 2.
For the DSC score, though the high DSC scores and low dispersed distributions from the violin plots of the liver segmentation indicate great performance, the results degrade for the other organs.
For the NSD score, the obtained scores and the dispersed distributions observed from the violin plots indicate unsatisfying segmentation performance for all four organs.
It is worth pointing out that for liver segmentation, the DSC scores are above 95\% for both subtasks, indicating great segmentation performance in terms of region overlap between the ground truth and the segmented region. NSD scores are 83\% and 85.8\% for the two subtasks respectively, demonstrating that the boundary regions contain more segmentation errors, which need further improvements. This phenomenon further proves the necessity of applying NSD for the evaluation of segmentation results.

Figure~\ref{fig:4-1-exp} shows segmentation results of some challenging examples from each subtask. It can be found that the SOTA method does not well generalize to lesion-affected organs. For example, the first row in Figure~\ref{fig:4-1-exp} shows a case with fatty liver in which the liver is darker than the healthy cases. The SOTA method fails to segment the liver completely. The spleen (blue) segmentation result is also poor in this situation. Moreover, the cases in the 2$^{nd}$, 3$^{rd}$, and 4$^{th}$ rows have kidney (green), and spleen (blue) tumors, respectively. There exist serious under-segmentation and incorrect segmentation in the segmentation results.
These challenging cases are still unsolved problems for abdominal organ segmentation, which are not highlighted in current publicly available benchmarks.

\begin{table}[!htbp]
\centering
\caption{Task settings of semi-supervised multi-organ segmentation.}
\label{tab:exp-semi}
\begin{tabular}{cc|c|c}
\hline
\multicolumn{2}{c|}{Training}                                                                                       & \multirow{2}{*}{Testing}    & \multirow{2}{*}{Note} \\ \cline{1-2}
Labelled  & Unlabelled & &   \\ \hline
Spl. Plus (41)  & -  & \multirow{11}{*}{\begin{tabular}[c]{@{}c@{}} 100\\ cases \end{tabular}}                       & Lower Bound           \\ \cline{1-2,4} 
Spl. Plus (41)                                                                                           & Pan. Plus (400)                                                                   &                                                                    & Subtask 1                     \\ \cline{1-2,4} 
Spl. Plus (41)                                                                                           & \begin{tabular}[c]{@{}c@{}} Pan. Plus (400)\\ LiTS Plus (145)\\ KiTS Plus (250)\\ Spl. Plus Ts (5)\end{tabular} &                                                                    & Subtask 2                      \\ \cline{1-2,4} 
\begin{tabular}[c]{@{}c@{}}Spl. Plus (41)\\ Pan. Plus (400)\\ LiTS Plus (145)\\ KiTS Plus (250)\\ Spl. Plus Ts (5)\end{tabular} & -                                                                                        &                                                                    & Upper Bound           \\ \hline
\end{tabular}
\end{table}


\subsection{Semi-supervised organ segmentation benchmark}
Semi-supervised learning is an effective way to utilize unlabelled data and reduce annotation demand, which is an active research topic currently. There are several benchmarks in the natural image/video segmentation domain~\cite{perazzi2016DAVIS, pont2017DAVIS}. However, there still exists no related benchmark in the medical image segmentation community. Thus, we set up this benchmark to explore how we can use the unlabelled data to boost the performance of abdominal organ segmentation.

\subsubsection{Task setting}
\textbf{Motivation of the training set and the testing set choice:}
this semi-supervised task employs MSD Spleen Plus, LiTS Plus, KiTS Plus, MSD Pancreas Plus, and NIH Pancreas Plus as the training and testing datasets.
The semi-supervised task is dedicated to alleviating the burden of manual annotations. In this scenario, a small portion of labelled data and a large amount of unlabelled data are available. Therefore, we set the smallest subset, Spleen Plus with 41 cases, as the labelled training set. To show the superiority of semi-supervised methods for leveraging a large amount of unlabelled data, approximately 10-20 times amount of data (400-800 cases) from the remaining subsets are selected as the unlabelled training set. We use the baseline model in Section~\ref{sec:semi:baseline} to infer all the remaining cases and select 100 cases as the final testing set, including 50 challenging cases with the lowest average DSC and NSD scores and 50 randomly selected cases.

Table~\ref{tab:exp-semi} presents the semi-supervised segmentation benchmark settings that consist of 2 subtasks. As a contrast, we start with a fully supervised lower-bound task, where a model is trained solely on MSD Spleen Plus containing 41 well-annotated cases. The upper-bound task is also fully supervised that involves the additional 800 labelled cases. Precisely, in upper-bound training set, 41 cases are from MSD Spleen Plus, 400 cases are from MSD and NIH Pancreas Plus, 145 cases are from LiTS Plus, 250 cases are from KiTS Plus, and 5 cases are from MSD Spleen Plus testing set. Based on the lower-bound and upper-bound subtasks, unlabelled cases are gradually involved in the following semi-supervised subtasks. In order to evaluate the effect of the unlabelled data and their quantity on multi-organ segmentation, we carefully design 2 subtasks concerning the source and quantity of unlabelled data. Specifically, subtask 1 utilizes 400 unlabelled cases from MSD and NIH Pancreas Plus, and in addition to the 400 cases, subtask 2 exploits additional 400 unlabelled cases from LiTS plus, KiTS plus, and MSD Spleen Plus testing set. Both subtasks are evaluated on the consistent hold-out testing set for fair comparisons.

\begin{table*}
\centering
\caption{Quantitative multi-organ segmentation results in semi-supervised benchmark.}
\label{tab:4-2-organ-detail}
\begin{tabular}{ccccccccccc}
\hline
\multirow{2}{*}{Task} & \multicolumn{2}{c}{Liver}       & \multicolumn{2}{c}{Kidney}      & \multicolumn{2}{c}{Spleen}      & \multicolumn{2}{c}{Pancreas} & \multicolumn{2}{c}{Average}    \\
\cline{2-11}
                      & DSC (\%)            & NSD (\%)            & DSC (\%)            & NSD (\%)            & DSC (\%)            & NSD (\%)            & DSC (\%)            & NSD (\%)            & DSC (\%)            & NSD (\%)             \\
\cline{2-11}
Lower Bound           & 95.7$\pm$5.3  & 83.0$\pm$11  & 91.3$\pm$14  & 83.8$\pm$12  & 93.6$\pm$12  & 88.2$\pm$15  & 81.5$\pm$15  & 67.7$\pm$16  & 90.5$\pm$13  & 80.7$\pm$16   \\
Subtask 1             & 96.2$\pm$4.2  & 84.0$\pm$9.7  & 91.5$\pm$12  & 83.6$\pm$12  & 94.6$\pm$11  & 90.4$\pm$14  & 82.8$\pm$13  & 69.2$\pm$16  & 91.3$\pm$12  & 81.8$\pm$15   \\
Subtask 2             & 96.2$\pm$4.0  & 83.7$\pm$9.5  & 92.2$\pm$12  & 84.3$\pm$11  & 94.9$\pm$11  & 90.6$\pm$13  & 82.9$\pm$13  & 68.4$\pm$15  & 91.5$\pm$12  & 81.8$\pm$15   \\
Upper Bound           &  \textbf{97.4$\pm$2.3}  & \textbf{86.7$\pm$8.6}  & \textbf{95.4$\pm$4.0}  & \textbf{86.9$\pm$8.3}  & \textbf{96.0$\pm$9.5}  & \textbf{92.9$\pm$12}  & \textbf{85.7$\pm$8.9}  & \textbf{72.5$\pm$13}  & \textbf{93.6$\pm$8.3}  & \textbf{84.7$\pm$13}   \\
\hline

\end{tabular}
\end{table*}

\begin{figure*}[!htbp]
\centering
\includegraphics[scale=0.45]{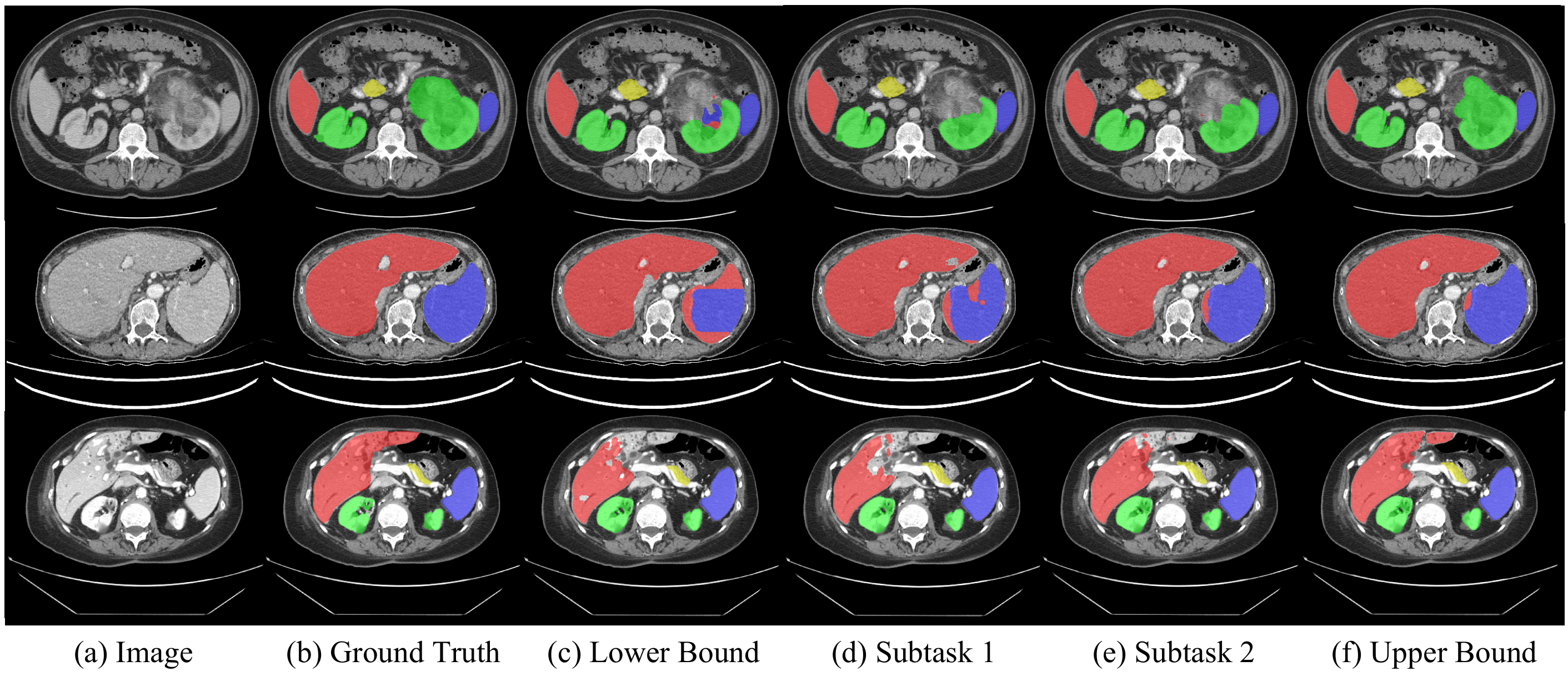}
\caption{Challenging examples from testing sets in semi-supervised segmentation benchmark.}
\label{fig:4-2-exp}
\end{figure*}

\subsubsection{Baseline and results}
\label{sec:semi:baseline}
Motivated by the success of the noisy-student learning method in semi-supervised image classification~\cite{CVPR20-noisy-student} and semi-supervised urban scene segmentation~\cite{ECCV20-noisy-student-seg} tasks, we develop a teacher-student-based method for semi-supervised abdominal organ segmentation, which includes five main steps:
\begin{itemize}
    \item Step 1. Training a teacher model on the manually labelled data.
    \item Step 2. Generating pseudo labels of the unlabelled data via the teacher model.
    \item Step 3. Training a student model on both manual and pseudo labelled data.
    \item Step 4. Finetuning the student model in step 3 on the manually labelled data.
    \item Step 5. Going back to step 2 and replacing the teacher model with the student model for a desired number of iterations.
\end{itemize}

In the experiments, we employ 3D nnU-Net for both teacher and student models. The results are presented in Table~\ref{tab:4-2-organ-detail}. Due to the different quantity of labelled cases during training, there exists a performance gap between the lower-bound and the upper-bound subtasks. With unlabelled data involved, the performance gradually increased in terms of the average DSC and NSD, indicating that the proposed method can leverage unlabelled cases to improve the multi-organ segmentation performance.

Figure~\ref{fig:4-2-exp} illustrates segmentation results of 3 challenging examples from each subtask. It is observed that our semi-supervised method is able to reduce misclassification by leveraging unlabelled data. 
The first and third rows show cases with a large kidney tumor and cholangiectasis inside the liver, respectively. The pathology changes pose an extreme challenge for the kidney segmentation. The second row demonstrates a case that the spleen shares similar appearances with the liver, where it tends to be recognized as the liver when the training data is limited. 
We can also find that the segmentation error can be gradually corrected by utilizing more unlabelled data. The violin plots of the segmentation results in Supplementary Figure 3 show that a performance increasing trend is observed for the four organs when the quantity of the unlabelled data is increased.


\subsection{Weakly supervised abdominal organ segmentation benchmark}
\label{ss:weakly}
This benchmark is to explore how we can use weak annotations to generate full segmentation results. There are several different weak annotation strategies for segmentation tasks, such as random scribbles, bounding boxes, extreme points and sparse labels.
Sparse labels are the most commonly used weak annotations for organs segmentation when radiologists manually delineate the organs \cite{KiTSDataset}. In this benchmark, we provide slice-level sparse labels in the training set, where only part ($\leq 30\%$) of the slices are well annotated.


\subsubsection{Task settings}
\textbf{Motivation of the training set and the testing set choice:} we select the Spleen Plus (41) as the training set because it has the least training cases. This choice is more in line with reality compared with using other datasets (e.g., KiTS Plus (210), LiTS Plus (131)), because the training set has only limited well-annotated cases in many medical centers. 

The weakly supervised organ segmentation benchmark contains three subtasks as shown in 
Table~\ref{tab:weaklySup}, in which only a fraction of the slices are annotated at roughly uniform intervals. We generate sparse labels with roughly uniform intervals because, in practice, human-raters usually annotate such sparse labels and then interpolate the unlabelled slices \cite{KiTSDataset}. 
Specifically, we set three different annotation rates 5\%, 15\%, and 30\%, which are similar to the existing work \cite{zhang2019sparse} on brain tissue segmentation. We use the baseline model in Section~\ref{sec:weak:baseline} to infer all the remaining cases and select 100 cases as the final testing set, including 50 challenging cases with the lowest average DSC and NSD scores and 50 randomly selected cases.

\begin{table}[!htbp]
\centering
\caption{Task settings and quantitative baseline results of weakly supervised abdominal organ segmentation.}\label{tab:weaklySup}
\renewcommand\tabcolsep{2pt}
\begin{tabular}{llccc}
\hline
\multicolumn{1}{c}{Training}                                                                    & \multicolumn{1}{c}{Ratio} & Testing                                                                                             & DSC (\%)                           & NSD (\%)                            \\
\hline
\multirow{3}{*}{\begin{tabular}[c]{@{}l@{}}Spleen Plus (41)\\ \end{tabular}} & 5\%                       & \multirow{3}{*}{\begin{tabular}[c]{@{}c@{}}100 cases \end{tabular}} & 78.0 $\pm$ 21.8                    & 63.5 $\pm$ 20.2                 \\
                                                                                                & 15\%                      &                                                                                                     & 83.9 $\pm$ 17.5                     & 70.4 $\pm$ 18.1                      \\
                                                                                                & 30\%                      &                                                                                                     & \multicolumn{1}{l}{\textbf{84.7 $\pm$ 16.7}} & \multicolumn{1}{l}{\textbf{70.9 $\pm$ 17.6}}  \\
\hline
\end{tabular}
\end{table}

\begin{table*}
\centering
\caption{Quantitative multi-organ segmentation results in weakly supervised benchmark.}
\label{tab:4-3-organ-details}
\begin{tabular}{cccccccccc}
\hline
\multirow{2}{*}{Task}        & \multirow{2}{*}{Method} & \multicolumn{2}{c}{Liver}                   & \multicolumn{2}{c}{Kidney}                  & \multicolumn{2}{c}{Spleen}                  & \multicolumn{2}{c}{Pancreas}                \\ \cline{3-10} 
                             &                         & DSC (\%)                  & NSD (\%)                  & DSC (\%)                  & NSD (\%)                  & DSC (\%)                  & NSD (\%)                  & DSC (\%)                  & NSD (\%)                  \\ \hline
\multirow{2}{*}{5\% labels}  & 2D U-Net                & 92.5 $\pm$ 6.50        & 72.7 $\pm$ 12.0       & \textbf{80.3 $\pm$ 19.2}        & \textbf{68.9 $\pm$ 15.1}        & \textbf{82.0 $\pm$ 21.7}        & 69.0 $\pm$ 22.5        & \textbf{57.2 $\pm$ 18.9}        & \textbf{43.3 $\pm$ 14.4}        \\
                             & 2D U-Net + CRF          & \multicolumn{1}{l}{\textbf{92.7 $\pm$ 6.19}} & \multicolumn{1}{l}{\textbf{72.9 $\pm$ 12.1}} & \multicolumn{1}{l}{78.3 $\pm$ 19.7} & \multicolumn{1}{l}{65.1 $\pm$ 15.2} & \multicolumn{1}{l}{81.8 $\pm$ 22.7} & \multicolumn{1}{l}{\textbf{70.3 $\pm$ 23.6}} & \multicolumn{1}{l}{55.2 $\pm$ 19.6} & \multicolumn{1}{l}{42.5 $\pm$ 15.2} \\ \hline
\multirow{2}{*}{15\% labels} & 2D U-Net                & 93.5 $\pm$ 6.15        & 76.5 $\pm$ 10.9        & \textbf{85.0 $\pm$ 17.1}        & \textbf{75.4 $\pm$ 15.1}        & 88.7 $\pm$ 15.7        & 76.6 $\pm$ 18.8        & \textbf{68.5 $\pm$ 17.5}        & \textbf{52.9 $\pm$ 14.8}        \\
                             & 2D U-Net + CRF          & \multicolumn{1}{l}{\textbf{93.7 $\pm$ 5.89}} & \multicolumn{1}{l}{\textbf{77.2 $\pm$ 10.7}} & \multicolumn{1}{l}{83.4 $\pm$ 17.5} & \multicolumn{1}{l}{71.3 $\pm$ 16.0} & \multicolumn{1}{l}{\textbf{89.0 $\pm$ 16.2}} & \multicolumn{1}{l}{\textbf{79.0 $\pm$ 18.8}} & \multicolumn{1}{l}{67.2 $\pm$ 18.6} & \multicolumn{1}{l}{52.5 $\pm$ 16.2} \\ \hline
\multirow{2}{*}{30\% labels} & 2D U-Net                & 93.6 $\pm$ 6.07        & 76.6 $\pm$ 11.1        & \textbf{86.0 $\pm$ 16.1}        & \textbf{75.8 $\pm$ 14.5}        & 88.8 $\pm$ 15.4        & 76.1 $\pm$ 19.0       & \textbf{70.5 $\pm$ 17.0}        & \textbf{55.0 $\pm$ 14.7}        \\
                             & 2D U-Net + CRF          & \multicolumn{1}{l}{\textbf{93.8 $\pm$ 5.76}} & \multicolumn{1}{l}{\textbf{76.9 $\pm$ 11.1}} & \multicolumn{1}{l}{84.3 $\pm$ 16.5} & \multicolumn{1}{l}{72.0 $\pm$ 15.3} & \multicolumn{1}{l}{\textbf{89.1 $\pm$ 15.9}} & \multicolumn{1}{l}{\textbf{78.4 $\pm$ 19.4}} & \multicolumn{1}{l}{69.3 $\pm$ 18.1} & \multicolumn{1}{l}{54.5 $\pm$ 16.1} \\ \hline
\end{tabular}
\end{table*}

\subsubsection{Baseline and results}
\label{sec:weak:baseline}
Our baseline method is built on the combination of 2D nnU-Net \cite{isensee2020nnunet} and fully connected Conditional Random Fields (CRF)~\cite{FullyCRFs}, which is motivated by the method proposed in~\cite{gao2016CNN-CRF} where the missing annotation challenge was addressed. 
The main idea in \cite{gao2016CNN-CRF} is to train a pixel-wise classification (segmentation) network with limited labelled images and then segment the unlabelled image to obtain initial segmentation results, followed by a refinement step with fully connected CRF. Fully connected CRF has been widely used in many segmentation tasks (e.g., liver and liver tumor segmentation~\cite{MICCAI16-CNN-CRF-Liver,Yao17-CRF-Liver}, brain tumor segmentation~\cite{MIA18-CNN-CRF-BrainTumor}), which could be an effective way to refine segmentation results. Our new baseline also follows this idea and has the following three main steps:
\begin{itemize}
    \item Step 1. Training a 2D U-Net~\cite{isensee2020nnunet} with the sparse labels;
    \item Step 2. Obtaining segmentation probability maps by inferring the testing cases;
    \item Step 3. Refining the segmentation results with fully connected CRF where the unary potential is the probability map and the pairwise potentials are three Gaussian-kernel potentials defined by the CT attenuation scores~\cite{FullyCRFs,gao2016CNN-CRF}.
\end{itemize}

\begin{table*}[!htbp]
\centering
\caption{Task settings and quantitative baseline results of continual learning.}
\label{tab:Continual-benchmark}
\begin{tabular}{cccccc}
\hline
\multicolumn{2}{c}{Training}       & \multicolumn{2}{c}{Testing}                                                                                                                                                                                   & \multirow{2}{*}{DSC (\%)}       & \multirow{2}{*}{NSD (\%)}        \\
\cline{1-4}
Dataset               & Annotation & Dataset                                                                                                     & Annotation                                                                                      &                            &                             \\
\hline
MSD Pancreas Ts (139) & Liver      & \multirow{4}{*}{\begin{tabular}[c]{@{}c@{}}100 cases \end{tabular}} & \multirow{4}{*}{\begin{tabular}[c]{@{}c@{}}Liver, kidney,\\spleen, and\\pancreas \end{tabular}} & \multirow{4}{*}{80.6$\pm$10.1} & \multirow{4}{*}{69.8$\pm$9.77}  \\
KiTS (210)            & Kidney     &                                                                                                             &                                                                                                 &                            &                             \\
Spleen (41)           & Spleen     &                                                                                                             &                                                                                                 &                            &                             \\
MSD Pancreas (281)    & Pancreas   &                                                                                                             &                                                                                                 &                            &                             \\
\hline
\end{tabular}
\end{table*}
Table~\ref{tab:weaklySup} presents the average DSC and NSD scores for the four organs, and Table~\ref{tab:4-3-organ-details} presents the detailed segmentation results for each organ. As expected, the higher annotation ratio the training cases have, the better segmentation performance the baseline method can achieve. With only 15\% annotations, the baseline can achieve an average DSC score over 90\% for liver  segmentation. The results could motivate us to employ deep learning-based strategies to reduce manual annotation efforts and time.
In Supplementary Figure 4, we show violin plots of the segmentation results with different annotation ratios. The performance gains from 15\% to 30\% annotation ratio are fewer than the gains from 5\% to 15\%, indicating that naively adding annotations cannot always bring linear performance improvements.

In addition, it can be found that using CRF does not bring remarkable performance improvements.  A similar phenomenon has also been found by the winner solution~\cite{BraTS18-1st} in the well-known brain tumor segmentation (BraTS) challenge 2018. Although the results are not promising as expected, they offer new opportunities and challenges for traditional energy-based segmentation methods. Specifically, given initial (inaccurate) CNN segmentation results, how or what kind of energy-based models can consistently improve the segmentation accuracy? All the related results, including trained models, the used CRF code and hyper-parameter settings, the segmentation results and its probability maps will be publicly available for future research along this direction.

\subsection{Continual learning benchmark for abdominal organ segmentation}
Continual learning has been a newly  emerging research topic and attracted significant attention~\cite{continualReview19}. The goal is to explore how we should augment the trained segmentation model to learn new tasks without forgetting the learned tasks. There are several terms for such tasks, e.g., continual learning, incremental learning, life-long learning or online learning. In this paper, we use continual learning to denote such tasks, which is widely used in existing literatures~\cite{continualReview19}.
In CVPR 2020, the first continual learning benchmark, to the best of our knowledge, is set up for image classification\footnote{https://sites.google.com/view/clvision2020/challenge}. However, there is still a lack of a public continual learning benchmark for medical image segmentation. Therefore, we set up a continual learning benchmark for abdominal organ segmentation and develop a baseline solution.



\subsubsection{Task setting}
\textbf{Motivation of the training set and the testing set choice:} the original single organ datasets, including KiTS (210), Spleen (41), and MSD Pancreas (281), are used as training sets. 
To evaluate the generalization ability of approaches, we choose MSD Pancreas Ts (139) as the training set with only liver annotation rather than LiTS (131) dataset, because the LiTS (131) is a multi-center dataset that would be better to serve as a testing set.
We use the baseline model in Section~\ref{sec:continual:baseline} to infer all the remaining cases and select 100 cases as the final testing set, including 50 challenging cases with the lowest average DSC and NSD scores and 50 randomly selected cases. 
As shown in Table~\ref{tab:Continual-benchmark}, the training set contains four datasets where only one organ is annotated in each dataset. 
Specifically, the labels of MSD Pancreas Ts (139), KiTS (210), Spleen (41), and MSD Pancreas Ts (139) are liver, kidney, spleen, and pancreas, respectively. 
In a word, this task requires building a multi-organ segmentation model with only single organ annotated training sets.
It also should be noted that one cannot access the previous tasks' dataset when switching to a new task. For example, if a kidney segmentation model has been built with the KiTS (210) dataset, this dataset will be not available when augmenting the model to segment the spleen with Spleen (41) dataset.

\subsubsection{Baseline and results}
\label{sec:continual:baseline}
Motivated by the well-known learning without forgetting~\cite{TPAMI17-LearnNoForget}, we develop an embarrassingly simple but effective continual learning method as the baseline, which contains the following four steps:
\begin{itemize}
    \item Step 1. Individually training a liver segmentation nnU-Net~\cite{isensee2020nnunet} model based on the MSD Pancreas Ts (139) dataset.
    \item Step 2. Using the trained liver segmentation model to infer KiTS (210) and obtain pseudo liver labels. Thus, each case in the KiTS (210) has both liver and kidney labels. Then, we use the new labels to train a nnU-Net model that can segment both liver and kidney.
    \item Step 3. Using the trained model in Step 2 to infer Spleen (41) and obtain both liver and kidney pseudo labels. Thus, each case in the Spleen (210) has liver, kidney, and spleen labels. Then, we use the new labels to train a nnU-Net model that can segment the three organs.
    \item Step 4. Using the trained model in Step 3 to infer MSD Pancreas (281) and obtain liver, kidney and spleen pseudo labels. Thus, each case in the MSD Pancreas (281) has liver, kidney, spleen, and pancreas labels. Finally, we can obtain the final multi-organ segmentation model by training a nnU-Net with the new labels.
\end{itemize}

\begin{table}[!htbp]
\caption{Quantitative multi-organ segmentation results of continual learning.}\label{tab:4-4-organ-details}
\centering
\begin{tabular}{ccc}
\hline
Organ    & DSC (\%)       & NSD (\%)        \\
\hline
Liver    & 94.7$\pm$7.99 & 81.7$\pm$14.0  \\
Kidney   & 79.4$\pm$18.9 & 73.6$\pm$16.5  \\
Spleen   & 83.8$\pm$23.2 & 72.9$\pm$24.2  \\
Pancreas & 64.7$\pm$21.6 & 51.1$\pm$16.3  \\
\hline
\end{tabular}
\end{table}

\begin{table*}[!htbp]
\caption{Quantitative results on the common testing set of the four benchmarks.}\label{tab:common-testing}
\centering
\resizebox{\textwidth}{!}{
\begin{tabular}{cccccccccccc}
\hline
\multicolumn{2}{c}{\multirow{2}{*}{Task}}                                                  & \multicolumn{2}{c}{Liver}                             & \multicolumn{2}{c}{Kidney}                            & \multicolumn{2}{c}{Spleen}                             & \multicolumn{2}{c}{Pancreas}                          & \multicolumn{2}{c}{Average}         \\ 
\cline{3-12}
\multicolumn{2}{c}{}                                                                       & DSC (\%)                       & NSD (\%)                       & DSC (\%)                       & NSD (\%)                       & DSC (\%)                       & NSD (\%)                       & DSC (\%)                      & NSD (\%)                       & DSC (\%)            & NSD (\%)             \\ 
\hline
\multirow{2}{*}{\begin{tabular}[c]{@{}c@{}}Fully\\ Supervised\end{tabular}}  & Subtask 1   & 95.9$\pm$5.4
                     & 87.5$\pm$8.7
                     & 94.8$\pm$6.5
                     & 89.2$\pm$12
                     & 86.3$\pm$18
                     & 78.2$\pm$22
                     & 76.3$\pm$24
                     & 65.1$\pm$22
                     & 88.3$\pm$17
          & 80.0$\pm$20
           \\
                                                                             & Subtask 2   & 97.5$\pm$2.5
           & 89.3$\pm$5.2
            & \textbf{97.4$\pm$3.8}
            & \textbf{95.7$\pm$6.6}
            & \textbf{97.3$\pm$5.4}
            &\textbf{ 94.5$\pm$10}
           & \textbf{82.5$\pm$19}
            & \textbf{71.8$\pm$18}
           & \textbf{93.6$\pm$12}
 & \textbf{87.8$\pm$15}
  \\ 
\hline
\multirow{2}{*}{\begin{tabular}[c]{@{}c@{}}Semi-\\ Supervised\end{tabular}} 
                                                                             & Subtask 1   & 96.9$\pm$1.4
            & 83.2$\pm$7.1
                     & 96.0$\pm$4.4
                    & 90.0$\pm$7.6
                     & 93.3$\pm$11
                     & 86.3$\pm$18
                     & 72.5$\pm$20
                    & 57.9$\pm$19
                     & 89.7$\pm$15
          & 79.4$\pm$19
           \\
                                                                             & Subtask 2   & \textbf{97.9$\pm$1.0}
            & \textbf{91.2$\pm$4.0}
                     & 97.1$\pm$4.3
                     & 93.4$\pm$6.8
                     & 97.2$\pm$3.9
                     & 94.1$\pm$9.7
                     & 82.3$\pm$12
                     & 70.7$\pm$13
                     & \textbf{93.6$\pm$9.4}
          & 87.4$\pm$13
           \\

\hline
\multirow{3}{*}{\begin{tabular}[c]{@{}c@{}}Weakly\\ Supervised\end{tabular}} & Subtask 1   & 84.8$\pm$9.8
                     & 55.0$\pm$10
                     & 73.7$\pm$24
                     & 56.1$\pm$21
                     & 63.8$\pm$34
                     & 55.5$\pm$28
                     & 16.8$\pm$19
                     & 14.8$\pm$16
                     & 60$\pm$35
          & 45.6$\pm$27
           \\
                                                                             & Subtask 2   & \multicolumn{1}{l}{84.8$\pm$9.7
} & \multicolumn{1}{l}{55.3$\pm$11
} & \multicolumn{1}{l}{81.6$\pm$17
} & \multicolumn{1}{l}{62.4$\pm$19
} & \multicolumn{1}{l}{68.7$\pm$31
} & \multicolumn{1}{l}{58.6$\pm$27
} & \multicolumn{1}{l}{29.8$\pm$21
} & \multicolumn{1}{l}{20.4$\pm$17
} & 66.2$\pm$30
          & 49.2$\pm$25
           \\
                                                                             & Subtask 3   & 84.8$\pm$9.4
                     & 55.4$\pm$11
                     & 83.1$\pm$11
                     & 62.7$\pm$17
                     & 68.8$\pm$29
                     & 57.5$\pm$26
                     & 30.6$\pm$22
                     & 21.4$\pm$18
                     & 66.8$\pm$29
          & 49.2$\pm$25
           \\ 
\hline
\multicolumn{2}{c}{Continual Learning}                                                     & 93.6$\pm$9.8
                     & 79.6$\pm$13
                     & 90.3$\pm$13
                     & 81.7$\pm$16
                     & 80.8$\pm$24
                     & 67.1$\pm$22
                     & 77.0$\pm$20
                     & 59.1$\pm$19
                     & 85.4$\pm$19
          & 71.9$\pm$20
           \\
\hline
\end{tabular}}
\end{table*}

\begin{figure}[!h]
\centering
\includegraphics[scale=0.5]{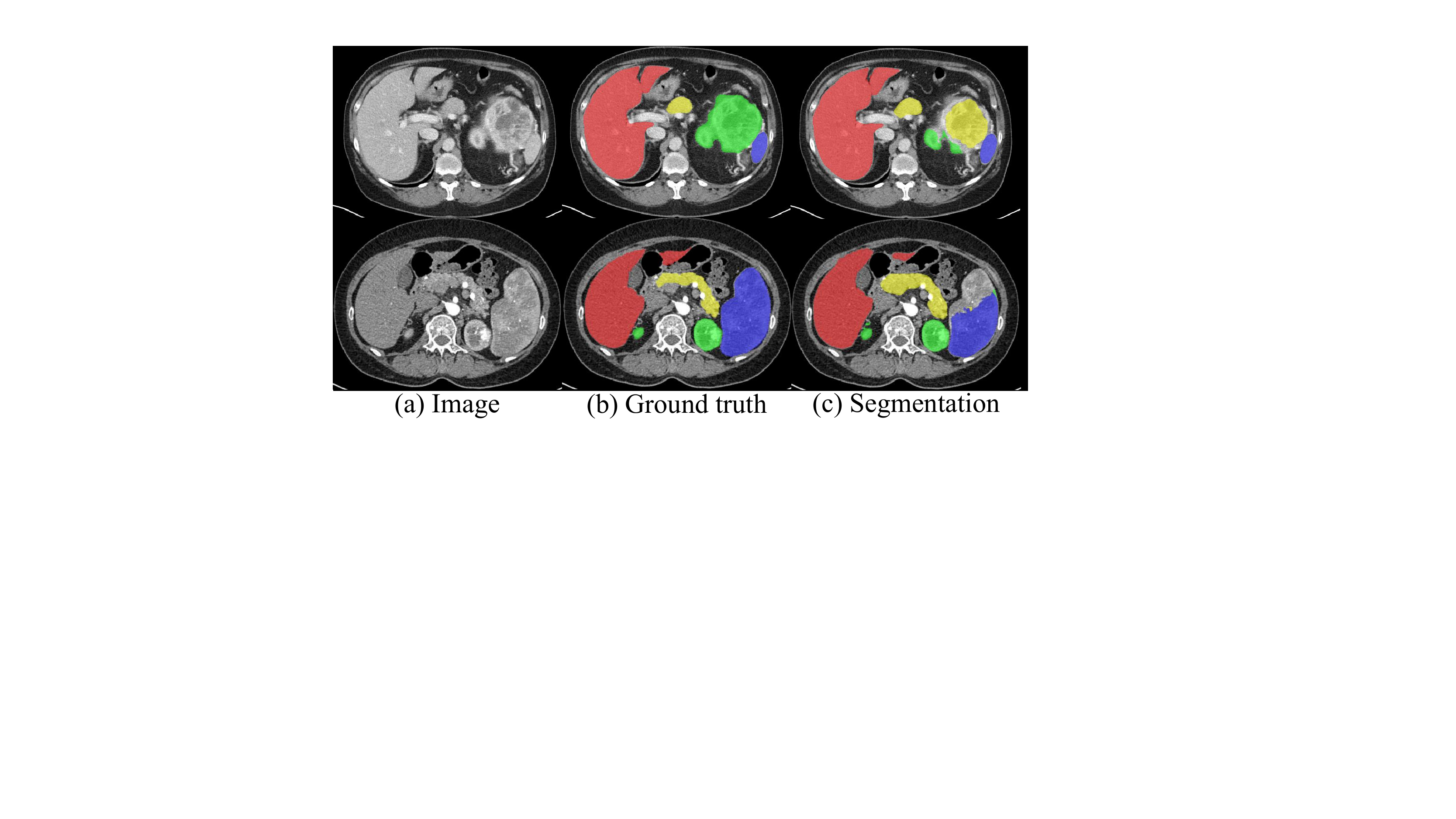}
\caption{Challenging examples from testing sets in continual learning multi-organ segmentation benchmark.}
\label{fig:4-4-examples}
\end{figure}

Table~\ref{tab:Continual-benchmark} presents the  average  DSC  and  NSD  scores  for  the  four organs,  and  Table~\ref{tab:4-4-organ-details}  presents  the  detailed  results  for  each organ. Overall, the performance of learning with single organ datasets is lower than learning with the full annotations as presented in the fully supervised segmentation results (Table~\ref{tab:exp-4organ}, Table~\ref{tab:fullSup-benchmark}), indicating that the model still tends to forget part of the previous tasks when switching to a new task.
The violin plots of the segmentation performance for each organ are presented in Supplementary Figure 5. Liver segmentation obtains the best DSC and NSD scores with compact distributions and fewer outliers while pancreas segmentation obtains lower performance.
The low scores and dispersed distributions of NSD reveal relatively high boundary errors because of the effects of various pathological changes as shown in Figure~\ref{fig:4-4-examples}.

\subsection{Evaluation and comparison on the common testing set}
The above testing sets are different in the four benchmarks. For an apple-to-apple comparison between the benchmarks, we introduce a common testing set (NJU dataset as described in Section~\ref{ss:dataset-motivation}) with 50 abdomen CT cases.



Table~\ref{tab:common-testing} presents the quantitative results of the common testing set of the four benchmarks. It is observed that the fully supervised method achieves the best average DSC and NSD scores for kidney, spleen and pancreas segmentation, because it uses many labelled cases. 
However, due to the burden of annotation, it is usually difficult to obtain the desired amount of annotations in clinical practice. Therefore, the problem lies in that what is the desired annotation-efficient method. The semi-supervised method in subtask 2 with only 41 labelled cases and 800 unlabelled cases achieves the best performance for the liver and the overall performance is very close to the best fully supervised method (361 labelled cases), indicating that using large and diverse unlabelled cases can significantly improve the performance. 
The weakly supervised methods achieve the lowest performance, but it requires the least annotation burden.

\section{Conclusion}
\label{S:conc}
In this work, we have introduced AbdomenCT-1K, the largest abdominal CT organ segmentation dataset, which includes multi-center, multi-phase, multi-vendor, and multi-disease cases. Although the SOTA method has achieved unprecedented performance in several existing benchmarks, such as liver, kidney, and spleen segmentation, our large-scale studies reveal that some problems remain unsolved as shown in Section~\ref{S:large-scale}.
In particular, the SOTA method can achieve superior segmentation results when the evaluation metric is DSC, the testing set has a similar data distribution as the training set, and no hard cases with unseen diseases in the testing set. However, the SOTA method cannot generalize the great performance on unseen datasets with many challenging cases, such as the cases with new CT phases, severe diseases, acquired from distinct scanners or clinical centers.

To advance the unsolved problems, we set up four new abdominal organ segmentation benchmarks, including fully supervised, semi-supervised, weakly supervised, and continual learning. Different from existing popular fully supervised abdominal organ segmentation benchmarks (e.g., LiTS \cite{bilic2019lits}, MSD \cite{simpson2019MSD}, and KiTS \cite{KiTS}), our new benchmarks have three main characteristics:
\begin{itemize}
    \item the testing cases in each benchmark are from multiple distinct CT scanners and medical centers.
    \item the challenging cases (e.g., with unseen or rare diseases) are selected and included in our testing sets, such as huge-tumor cases.
    \item instead of only focusing on the region-based metric (DSC), we also emphasize the boundary-related metric (NSD), because the boundary errors are critical in the preoperative planning of many abdominal organ surgeries, such as tumor resections and organ transplantation.
\end{itemize}

The main limitation is that we only focus on the segmentation of four large abdominal organs. However, there exist far more difficult organs~\cite{zhou2019semi,xie2019TinyOrgan} and lesions where the annotations are not available in our dataset. To address this limitation, we annotate 50 cases with 8 extra organs, including esophagus, gallbladder, stomach, aorta, celiac trunk, inferior vena cava, right adrenal gland and left adrenal gland. For the lesions, the detailed pathology information is not available in the original dataset. It is challenging to make a definite and accurate diagnosis with only CT scans because identifying the (malignant) tumor usually requires pathological examinations. As an alternative, we include the original single-organ tumor masks~\cite{bilic2019lits, simpson2019MSD, KiTS} and provide pseudo tumor labels of 663 cases by annotating all the other possible tumors, which can be used for noisy label learning.

Deep learning-based segmentation methods have achieved a great streak of successes. We hope that our large and diverse dataset and out-of-the-box baseline methods help push abdominal organ segmentation towards the real clinical practice.
\bibliographystyle{IEEEtran}
\bibliography{refs}

\begin{thebibliography}{100}
\providecommand{\url}[1]{#1}
\csname url@samestyle\endcsname
\providecommand{\newblock}{\relax}
\providecommand{\bibinfo}[2]{#2}
\providecommand{\BIBentrySTDinterwordspacing}{\spaceskip=0pt\relax}
\providecommand{\BIBentryALTinterwordstretchfactor}{4}
\providecommand{\BIBentryALTinterwordspacing}{\spaceskip=\fontdimen2\font plus
\BIBentryALTinterwordstretchfactor\fontdimen3\font minus
  \fontdimen4\font\relax}
\providecommand{\BIBforeignlanguage}[2]{{%
\expandafter\ifx\csname l@#1\endcsname\relax
\typeout{** WARNING: IEEEtran.bst: No hyphenation pattern has been}%
\typeout{** loaded for the language `#1'. Using the pattern for}%
\typeout{** the default language instead.}%
\else
\language=\csname l@#1\endcsname
\fi
#2}}
\providecommand{\BIBdecl}{\relax}
\BIBdecl

\bibitem{LiverSegPK09}
T.~Heimann, B.~Van~Ginneken, M.~A. Styner, Y.~Arzhaeva, V.~Aurich, C.~Bauer,
  A.~Beck, C.~Becker, R.~Beichel, G.~Bekes \emph{et~al.}, ``Comparison and
  evaluation of methods for liver segmentation from ct datasets,'' \emph{IEEE
  Transactions on Medical Imaging}, vol.~28, no.~8, pp. 1251--1265, 2009.

\bibitem{van2011computer}
B.~Van~Ginneken, C.~M. Schaefer-Prokop, and M.~Prokop, ``Computer-aided
  diagnosis: how to move from the laboratory to the clinic,'' \emph{Radiology},
  vol. 261, no.~3, pp. 719--732, 2011.

\bibitem{sykes2014reflections}
J.~Sykes, ``Reflections on the current status of commercial automated
  segmentation systems in clinical practice,'' \emph{Journal of Medical
  Radiation Sciences}, vol.~61, no.~3, pp. 131--134, 2014.

\bibitem{wang2019abdominal}
Y.~Wang, Y.~Zhou, W.~Shen, S.~Park, E.~K. Fishman, and A.~L. Yuille,
  ``Abdominal multi-organ segmentation with organ-attention networks and
  statistical fusion,'' \emph{Medical Image Analysis}, vol.~55, pp. 88--102,
  2019.

\bibitem{rater-variability}
L.~Joskowicz, D.~Cohen, N.~Caplan, and J.~Sosna, ``Inter-observer variability
  of manual contour delineation of structures in ct,'' \emph{European
  Radiology}, vol.~29, no.~3, pp. 1391--1399, 2019.

\bibitem{zhou2020review}
S.~K. Zhou, H.~Greenspan, C.~Davatzikos, J.~S. Duncan, B.~van Ginneken,
  A.~Madabhushi, J.~L. Prince, D.~Rueckert, and R.~M. Summers, ``A review of
  deep learning in medical imaging: Image traits, technology trends, case
  studies with progress highlights, and future promises,'' \emph{arXiv preprint
  arXiv:2008.09104}, 2020.

\bibitem{humpire2020fully}
G.~E. Humpire-Mamani, J.~Bukala, E.~T. Scholten, M.~Prokop, B.~van Ginneken,
  and C.~Jacobs, ``Fully automatic volume measurement of the spleen at ct using
  deep learning,'' \emph{Radiology: Artificial Intelligence}, vol.~2, no.~4, p.
  e190102, 2020.

\bibitem{RSNA-Checklist}
J.~Mongan, L.~Moy, and C.~E. Kahn, ``Checklist for artificial intelligence in
  medical imaging (claim): A guide for authors and reviewers,''
  \emph{Radiology: Artificial Intelligence}, vol.~2, no.~2, p. e200029, 2020.

\bibitem{NatureMedcine-Checklist}
B.~Norgeot, G.~Quer, B.~K. Beaulieu-Jones, A.~Torkamani, R.~Dias,
  M.~Gianfrancesco, R.~Arnaout, I.~S. Kohane, S.~Saria, E.~Topol \emph{et~al.},
  ``Minimum information about clinical artificial intelligence modeling: the
  mi-claim checklist,'' \emph{Nature Medicine}, vol.~26, no.~9, pp. 1320--1324,
  2020.

\bibitem{cheplygina2019not}
V.~Cheplygina, M.~de~Bruijne, and J.~P. Pluim, ``Not-so-supervised: a survey of
  semi-supervised, multi-instance, and transfer learning in medical image
  analysis,'' \emph{Medical Image Analysis}, vol.~54, pp. 280--296, 2019.

\bibitem{tajbakhsh2020embracing}
N.~Tajbakhsh, L.~Jeyaseelan, Q.~Li, J.~N. Chiang, Z.~Wu, and X.~Ding,
  ``Embracing imperfect datasets: A review of deep learning solutions for
  medical image segmentation,'' \emph{Medical Image Analysis}, p. 101693, 2020.

\bibitem{zhou2019prior}
Y.~Zhou, Z.~Li, S.~Bai, C.~Wang, X.~Chen, M.~Han, E.~Fishman, and A.~L. Yuille,
  ``Prior-aware neural network for partially-supervised multi-organ
  segmentation,'' in \emph{Proceedings of the IEEE International Conference on
  Computer Vision}, 2019, pp. 10\,672--10\,681.

\bibitem{shi2020marginal}
G.~Shi, L.~Xiao, Y.~Chen, and S.~K. Zhou, ``Marginal loss and exclusion loss
  for partially supervised multi-organ segmentation,'' \emph{Medical Image
  Analysis}, vol.~70, p. 101979, 2021.

\bibitem{DAVIS2020}
``{DAVIS: Densely Annotated VIdeo Segmentation},''
  \url{https://davischallenge.org/}, 2020, [Online; Accessed: Aug. 2020].

\bibitem{caelles20192019}
S.~Caelles, J.~Pont-Tuset, F.~Perazzi, A.~Montes, K.-K. Maninis, and
  L.~Van~Gool, ``The 2019 davis challenge on vos: Unsupervised multi-object
  segmentation,'' \emph{arXiv preprint arXiv:1905.00737}, 2019.

\bibitem{bilic2019lits}
P.~Bilic, P.~F. Christ, E.~Vorontsov, G.~Chlebus, H.~Chen, Q.~Dou, C.-W. Fu,
  X.~Han, P.-A. Heng, J.~Hesser \emph{et~al.}, ``The liver tumor segmentation
  benchmark (lits),'' \emph{arXiv preprint arXiv:1901.04056}, 2019.

\bibitem{KiTS}
N.~Heller, F.~Isensee, K.~H. Maier-Hein, X.~Hou, C.~Xie, F.~Li, Y.~Nan, G.~Mu,
  Z.~Lin, M.~Han \emph{et~al.}, ``The state of the art in kidney and kidney
  tumor segmentation in contrast-enhanced ct imaging: Results of the kits19
  challenge,'' \emph{Medical Image Analysis}, vol.~67, p. 101821, 2021.

\bibitem{meinzer2002LiverClinical}
H.-P. Meinzer, M.~Thorn, and C.~E. C{\'a}rdenas, ``Computerized planning of
  liver surgery—an overview,'' \emph{Computers \& Graphics}, vol.~26, no.~4,
  pp. 569--576, 2002.

\bibitem{ni2020LiverResection}
Z.-K. Ni, D.~Lin, Z.-Q. Wang, H.-M. Jin, X.-W. Li, Y.~Li, and H.~Huang,
  ``Precision liver resection: Three-dimensional reconstruction combined with
  fluorescence laparoscopic imaging,'' \emph{Surgical Innovation}, 2020.

\bibitem{simpson2019MSD}
A.~L. Simpson, M.~Antonelli, S.~Bakas, M.~Bilello, K.~Farahani,
  B.~Van~Ginneken, A.~Kopp-Schneider, B.~A. Landman, G.~Litjens, B.~Menze
  \emph{et~al.}, ``A large annotated medical image dataset for the development
  and evaluation of segmentation algorithms,'' \emph{arXiv preprint
  arXiv:1902.09063}, 2019.

\bibitem{NIHPancreas}
H.~R. Roth, A.~Farag, E.~B. Turkbey, L.~Lu, J.~Liu, and R.~M. Summers, ``Data
  from pancreas-ct,'' The Cancer Imaging Archive, 2016.

\bibitem{NIH-Pancreas2}
H.~R. Roth, L.~Lu, A.~Farag, H.-C. Shin, J.~Liu, E.~B. Turkbey, and R.~M.
  Summers, ``Deeporgan: Multi-level deep convolutional networks for automated
  pancreas segmentation,'' in \emph{International Conference on Medical Image
  Computing and Computer-assisted Intervention}, 2015, pp. 556--564.

\bibitem{TCIA}
K.~Clark, B.~Vendt, K.~Smith, J.~Freymann, J.~Kirby, P.~Koppel, S.~Moore,
  S.~Phillips, D.~Maffitt, M.~Pringle \emph{et~al.}, ``The cancer imaging
  archive (tcia): maintaining and operating a public information repository,''
  \emph{Journal of Digital Imaging}, vol.~26, no.~6, pp. 1045--1057, 2013.

\bibitem{isensee2020nnunet}
F.~Isensee, P.~F. J{\"a}eger, S.~A.~A. Kohl, J.~Petersen, and K.~H. Maier-Hein,
  ``nnu-net: a self-configuring method for deep learning-based biomedical image
  segmentation,'' \emph{Nature Methods}, vol.~18, no.~2, pp. 203--211, 2021.

\bibitem{nikolov2018SDice}
S.~Nikolov, S.~Blackwell, R.~Mendes, J.~De~Fauw, C.~Meyer, C.~Hughes,
  H.~Askham, B.~Romera-Paredes, A.~Karthikesalingam, C.~Chu \emph{et~al.},
  ``Deep learning to achieve clinically applicable segmentation of head and
  neck anatomy for radiotherapy,'' \emph{arXiv preprint arXiv:1809.04430},
  2018.

\bibitem{kass1988snakes}
M.~Kass, A.~Witkin, and D.~Terzopoulos, ``Snakes: Active contour models,''
  \emph{International Journal of Computer Vision}, vol.~1, no.~4, pp. 321--331,
  1988.

\bibitem{cootes1995ASM}
T.~F. Cootes, C.~J. Taylor, D.~H. Cooper, and J.~Graham, ``Active shape
  models-their training and application,'' \emph{Computer vision and image
  understanding}, vol.~61, no.~1, pp. 38--59, 1995.

\bibitem{iglesias2015multi}
J.~E. Iglesias and M.~R. Sabuncu, ``Multi-atlas segmentation of biomedical
  images: a survey,'' \emph{Medical Image Analysis}, vol.~24, no.~1, pp.
  205--219, 2015.

\bibitem{li2015automatic}
G.~Li, X.~Chen, F.~Shi, W.~Zhu, J.~Tian, and D.~Xiang, ``Automatic liver
  segmentation based on shape constraints and deformable graph cut in ct
  images,'' \emph{IEEE Transactions on Image Processing}, vol.~24, no.~12, pp.
  5315--5329, 2015.

\bibitem{HuPeijun-MP}
J.~Peng, P.~Hu, F.~Lu, Z.~Peng, D.~Kong, and H.~Zhang, ``3d liver segmentation
  using multiple region appearances and graph cuts,'' \emph{Medical Physics},
  vol.~42, no.~12, pp. 6840--6852, 2015.

\bibitem{CV-SegLiver}
S.~K. Siri and M.~V. Latte, ``Combined endeavor of neutrosophic set and
  chan-vese model to extract accurate liver image from ct scan,''
  \emph{Computer Methods and Programs in Biomedicine}, vol. 151, pp. 101--109,
  2017.

\bibitem{LiverSeg-SSM}
X.~Zhang, J.~Tian, K.~Deng, Y.~Wu, and X.~Li, ``Automatic liver segmentation
  using a statistical shape model with optimal surface detection,'' \emph{IEEE
  Transactions on Biomedical Engineering}, vol.~57, no.~10, pp. 2622--2626,
  2010.

\bibitem{zhou2005-atlas}
X.~Zhou, T.~Kitagawa, K.~Okuo, T.~Hara, H.~Fujita, R.~Yokoyama, M.~Kanematsu,
  and H.~Hoshi, ``Construction of a probabilistic atlas for automated liver
  segmentation in non-contrast torso ct images,'' in \emph{International
  Congress Series}, vol. 1281, 2005, pp. 1169--1174.

\bibitem{okada2015abdominal}
T.~Okada, M.~G. Linguraru, M.~Hori, R.~M. Summers, N.~Tomiyama, and Y.~Sato,
  ``Abdominal multi-organ segmentation from ct images using conditional
  shape--location and unsupervised intensity priors,'' \emph{Medical Image
  Analysis}, vol.~26, no.~1, pp. 1--18, 2015.

\bibitem{xu2015efficient}
Z.~Xu, R.~P. Burke, C.~P. Lee, R.~B. Baucom, B.~K. Poulose, R.~G. Abramson, and
  B.~A. Landman, ``Efficient multi-atlas abdominal segmentation on clinically
  acquired ct with simple context learning,'' \emph{Medical Image Analysis},
  vol.~24, no.~1, pp. 18--27, 2015.

\bibitem{ronneberger20152DUNet}
O.~Ronneberger, P.~Fischer, and T.~Brox, ``U-net: convolutional networks for
  biomedical image segmentation,'' in \emph{International Conference on Medical
  Image Computing and Computer-assisted Intervention}, 2015, pp. 234--241.

\bibitem{ACDC-TMI}
O.~Bernard, A.~Lalande, C.~Zotti, F.~Cervenansky, X.~Yang, P.-A. Heng,
  I.~Cetin, K.~Lekadir, O.~Camara, M.~A.~G. Ballester \emph{et~al.}, ``Deep
  learning techniques for automatic mri cardiac multi-structures segmentation
  and diagnosis: is the problem solved?'' \emph{IEEE Transactions on Medical
  Imaging}, vol.~37, no.~11, pp. 2514--2525, 2018.

\bibitem{BRATS-Review}
S.~Bakas, M.~Reyes, A.~Jakab, S.~Bauer, M.~Rempfler, A.~Crimi, R.~T. Shinohara,
  C.~Berger, S.~M. Ha, M.~Rozycki \emph{et~al.}, ``Identifying the best machine
  learning algorithms for brain tumor segmentation, progression assessment, and
  overall survival prediction in the brats challenge,'' \emph{arXiv preprint
  arXiv:1811.02629}, 2018.

\bibitem{seo2019modified}
H.~Seo, C.~Huang, M.~Bassenne, R.~Xiao, and L.~Xing, ``Modified u-net (mu-net)
  with incorporation of object-dependent high level features for improved liver
  and liver-tumor segmentation in ct images,'' \emph{IEEE Transactions on
  Medical Imaging}, vol.~39, no.~5, pp. 1316--1325, 2019.

\bibitem{MICCAI2019-shape-prior-liver-seg}
J.~Yao, J.~Cai, D.~Yang, D.~Xu, and J.~Huang, ``Integrating 3d geometry of
  organ for improving medical image segmentation,'' in \emph{International
  Conference on Medical Image Computing and Computer-Assisted Intervention},
  2019, pp. 318--326.

\bibitem{DLMI2017-PHNN}
K.~George, A.~P. Harrison, D.~Jin, Z.~Xu, and D.~J. Mollura, ``Pathological
  pulmonary lobe segmentation from ct images using progressive holistically
  nested neural networks and random walker,'' in \emph{Deep learning in Medical
  Image Analysis and Multimodal Learning for Clinical Decision Support}, 2017,
  pp. 195--203.

\bibitem{MICCAI2017-PHNN}
A.~P. Harrison, Z.~Xu, K.~George, L.~Lu, R.~M. Summers, and D.~J. Mollura,
  ``Progressive and multi-path holistically nested neural networks for
  pathological lung segmentation from ct images,'' in \emph{International
  Conference on Medical Image Computing and Computer Assisted Intervention},
  2017, pp. 621--629.

\bibitem{mia2021-deeptarget}
D.~Jin, D.~Guo, T.-Y. Ho, A.~P. Harrison, J.~Xiao, C.-k. Tseng, and L.~Lu,
  ``Deeptarget: Gross tumor and clinical target volume segmentation in
  esophageal cancer radiotherapy,'' \emph{Medical Image Analysis}, vol.~68, p.
  101909, 2021.

\bibitem{zhou2017fixed}
Y.~Zhou, L.~Xie, W.~Shen, Y.~Wang, E.~K. Fishman, and A.~L. Yuille, ``A
  fixed-point model for pancreas segmentation in abdominal ct scans,'' in
  \emph{International Conference on Medical Image Computing and
  Computer-assisted Intervention}, 2017, pp. 693--701.

\bibitem{mia2018-spatial-aggre-pancreas-seg}
H.~R. Roth, L.~Lu, N.~Lay, A.~P. Harrison, A.~Farag, A.~Sohn, and R.~M.
  Summers, ``Spatial aggregation of holistically-nested convolutional neural
  networks for automated pancreas localization and segmentation,''
  \emph{Medical Image Analysis}, vol.~45, pp. 94--107, 2018.

\bibitem{roth2016spatial}
H.~R. Roth, L.~Lu, A.~Farag, A.~Sohn, and R.~M. Summers, ``Spatial aggregation
  of holistically-nested networks for automated pancreas segmentation,'' in
  \emph{International Conference on Medical Image Computing and
  Computer-assisted Intervention}, 2016, pp. 451--459.

\bibitem{xue2019cascaded}
J.~Xue, K.~He, D.~Nie, E.~Adeli, Z.~Shi, S.-W. Lee, Y.~Zheng, X.~Liu, D.~Li,
  and D.~Shen, ``Cascaded multitask 3-d fully convolutional networks for
  pancreas segmentation,'' \emph{IEEE Transactions on Cybernetics}, 2019.

\bibitem{TMI-LGAC}
M.~Jun, H.~Jian, and Y.~Xiaoping, ``Learning geodesic active contours for
  embedding object global information in segmentation cnns,'' \emph{IEEE
  Transactions on Medical Imaging}, 2020.

\bibitem{3DV2019-v-nas}
Z.~Zhu, C.~Liu, D.~Yang, A.~Yuille, and D.~Xu, ``V-nas: Neural architecture
  search for volumetric medical image segmentation,'' in \emph{2019
  International Conference on 3D Vision}, 2019, pp. 240--248.

\bibitem{2020cvpr-organ-at-risk-nas}
D.~Guo, D.~Jin, Z.~Zhu, T.-Y. Ho, A.~P. Harrison, C.-H. Chao, J.~Xiao, and
  L.~Lu, ``Organ at risk segmentation for head and neck cancer using stratified
  learning and neural architecture search,'' in \emph{Proceedings of the
  IEEE/CVF Conference on Computer Vision and Pattern Recognition}, 2020, pp.
  4223--4232.

\bibitem{2021cvpr-nas-3d-medical}
Y.~He, D.~Yang, H.~Roth, C.~Zhao, and D.~Xu, ``Dints: Differentiable neural
  network topology search for 3d medical image segmentation,'' in \emph{IEEE
  Conference on Computer Vision and Pattern Recognition}, 2021.

\bibitem{roth2018application}
H.~R. Roth, H.~Oda, X.~Zhou, N.~Shimizu, Y.~Yang, Y.~Hayashi, M.~Oda,
  M.~Fujiwara, K.~Misawa, and K.~Mori, ``An application of cascaded 3d fully
  convolutional networks for medical image segmentation,'' \emph{Computerized
  Medical Imaging and Graphics}, vol.~66, pp. 90--99, 2018.

\bibitem{gibson2018automatic}
E.~Gibson, F.~Giganti, Y.~Hu, E.~Bonmati, S.~Bandula, K.~Gurusamy, B.~Davidson,
  S.~P. Pereira, M.~J. Clarkson, and D.~C. Barratt, ``Automatic multi-organ
  segmentation on abdominal ct with dense v-networks,'' \emph{IEEE Transactions
  on Medical Imaging}, vol.~37, no.~8, pp. 1822--1834, 2018.

\bibitem{zhou2016three}
X.~Zhou, T.~Ito, R.~Takayama, S.~Wang, T.~Hara, and H.~Fujita,
  ``Three-dimensional ct image segmentation by combining 2d fully convolutional
  network with 3d majority voting,'' in \emph{Deep Learning and Data Labeling
  for Medical Applications}, 2016, pp. 111--120.

\bibitem{larsson2018robust}
M.~Larsson, Y.~Zhang, and F.~Kahl, ``Robust abdominal organ segmentation using
  regional convolutional neural networks,'' \emph{Applied Soft Computing},
  vol.~70, pp. 465--471, 2018.

\bibitem{hu2017automatic}
P.~Hu, F.~Wu, J.~Peng, Y.~Bao, F.~Chen, and D.~Kong, ``Automatic abdominal
  multi-organ segmentation using deep convolutional neural network and
  time-implicit level sets,'' \emph{International Journal of Computer Assisted
  Radiology and Surgery}, vol.~12, no.~3, pp. 399--411, 2017.

\bibitem{roth2018multi}
H.~R. Roth, C.~Shen, H.~Oda, T.~Sugino, M.~Oda, Y.~Hayashi, K.~Misawa, and
  K.~Mori, ``A multi-scale pyramid of 3d fully convolutional networks for
  abdominal multi-organ segmentation,'' in \emph{International Conference on
  Medical Image Computing and Computer-assisted Intervention}, 2018, pp.
  417--425.

\bibitem{zhang2020block}
L.~Zhang, J.~Zhang, P.~Shen, G.~Zhu, P.~Li, X.~Lu, H.~Zhang, S.~A. Shah, and
  M.~Bennamoun, ``Block level skip connections across cascaded v-net for
  multi-organ segmentation,'' \emph{IEEE Transactions on Medical Imaging},
  2020.

\bibitem{milletari2016v}
F.~Milletari, N.~Navab, and S.-A. Ahmadi, ``V-net: Fully convolutional neural
  networks for volumetric medical image segmentation,'' in \emph{Fourth
  International Conference on 3D vision}, 2016, pp. 565--571.

\bibitem{heinrich2019obelisk}
M.~P. Heinrich, O.~Oktay, and N.~Bouteldja, ``Obelisk-net: Fewer layers to
  solve 3d multi-organ segmentation with sparse deformable convolutions,''
  \emph{Medical Image Analysis}, vol.~54, pp. 1--9, 2019.

\bibitem{van2020survey}
J.~E. Van~Engelen and H.~H. Hoos, ``A survey on semi-supervised learning,''
  \emph{Machine Learning}, vol. 109, no.~2, pp. 373--440, 2020.

\bibitem{lee2013pseudo}
D.-H. Lee, ``Pseudo-label: The simple and efficient semi-supervised learning
  method for deep neural networks,'' in \emph{Workshop on Challenges in
  Representation Learning, ICML}, vol.~3, 2013, p.~2.

\bibitem{iscen2019label}
A.~Iscen, G.~Tolias, Y.~Avrithis, and O.~Chum, ``Label propagation for deep
  semi-supervised learning,'' in \emph{Proceedings of the IEEE Conference on
  Computer Vision and Pattern Recognition}, 2019, pp. 5070--5079.

\bibitem{zhou2019semi}
Y.~Zhou, Y.~Wang, P.~Tang, S.~Bai, W.~Shen, E.~Fishman, and A.~Yuille,
  ``Semi-supervised 3d abdominal multi-organ segmentation via deep multi-planar
  co-training,'' in \emph{2019 IEEE Winter Conference on Applications of
  Computer Vision}, 2019, pp. 121--140.

\bibitem{lee2020semi}
H.~H. Lee, Y.~Tang, O.~Tang, Y.~Xu, Y.~Chen, D.~Gao, S.~Han, R.~Gao, M.~R.
  Savona, R.~G. Abramson \emph{et~al.}, ``Semi-supervised multi-organ
  segmentation through quality assurance supervision,'' in \emph{Medical
  Imaging 2020: Image Processing}, vol. 11313, 2020, p. 113131I.

\bibitem{ECCV2020-semi-liver}
A.~Raju, C.-T. Cheng, Y.~Huo, J.~Cai, J.~Huang, J.~Xiao, L.~Lu, C.~Liao, and
  A.~P. Harrison, ``Co-heterogeneous and adaptive segmentation from
  multi-source and multi-phase ct imaging data: a study on pathological liver
  and lesion segmentation,'' in \emph{European Conference on Computer Vision},
  2020, pp. 448--465.

\bibitem{mia2020-semi-uncertainty}
Y.~Xia, D.~Yang, Z.~Yu, F.~Liu, J.~Cai, L.~Yu, Z.~Zhu, D.~Xu, A.~Yuille, and
  H.~Roth, ``Uncertainty-aware multi-view co-training for semi-supervised
  medical image segmentation and domain adaptation,'' \emph{Medical Image
  Analysis}, vol.~65, p. 101766, 2020.

\bibitem{kanavati2017joint}
F.~Kanavati, K.~Misawa, M.~Fujiwara, K.~Mori, D.~Rueckert, and B.~Glocker,
  ``Joint supervoxel classification forest for weakly-supervised organ
  segmentation,'' in \emph{International Workshop on Machine Learning in
  Medical Imaging}, 2017, pp. 79--87.

\bibitem{zeng2019weakly}
H.~Zeng, X.~Hu, L.~Chen, C.~Zhou, and Y.~Wen, ``Weakly supervised learning of
  recurrent residual convnets for pancreas segmentation in ct scans,'' in
  \emph{2019 IEEE International Conference on Bioinformatics and Biomedicine},
  2019, pp. 1409--1415.

\bibitem{song2019box}
C.~Song, Y.~Huang, W.~Ouyang, and L.~Wang, ``Box-driven class-wise region
  masking and filling rate guided loss for weakly supervised semantic
  segmentation,'' in \emph{Proceedings of the IEEE Conference on Computer
  Vision and Pattern Recognition}, 2019, pp. 3136--3145.

\bibitem{bearman2016s}
A.~Bearman, O.~Russakovsky, V.~Ferrari, and L.~Fei-Fei, ``What’s the point:
  Semantic segmentation with point supervision,'' in \emph{European Conference
  on Computer Vision}, 2016, pp. 549--565.

\bibitem{qian2019weakly}
R.~Qian, Y.~Wei, H.~Shi, J.~Li, J.~Liu, and T.~Huang, ``Weakly supervised scene
  parsing with point-based distance metric learning,'' in \emph{Proceedings of
  the AAAI Conference on Artificial Intelligence}, vol.~33, 2019, pp.
  8843--8850.

\bibitem{lin2016scribblesup}
D.~Lin, J.~Dai, J.~Jia, K.~He, and J.~Sun, ``Scribblesup: Scribble-supervised
  convolutional networks for semantic segmentation,'' in \emph{Proceedings of
  the IEEE Conference on Computer Vision and Pattern Recognition}, 2016, pp.
  3159--3167.

\bibitem{ji2019scribble}
Z.~Ji, Y.~Shen, C.~Ma, and M.~Gao, ``Scribble-based hierarchical weakly
  supervised learning for brain tumor segmentation,'' in \emph{International
  Conference on Medical Image Computing and Computer-Assisted Intervention},
  2019, pp. 175--183.

\bibitem{pathak2015constrained}
D.~Pathak, P.~Krahenbuhl, and T.~Darrell, ``Constrained convolutional neural
  networks for weakly supervised segmentation,'' in \emph{Proceedings of the
  IEEE International Conference on Computer Vision}, 2015, pp. 1796--1804.

\bibitem{liu2020leveraging}
Y.~Liu, Y.-H. Wu, P.-S. Wen, Y.-J. Shi, Y.~Qiu, and M.-M. Cheng, ``Leveraging
  instance-, image-and dataset-level information for weakly supervised instance
  segmentation,'' \emph{IEEE Transactions on Pattern Analysis and Machine
  Intelligence}, 2020.

\bibitem{wang2020weakly}
X.~Wang, S.~Liu, H.~Ma, and M.-H. Yang, ``Weakly-supervised semantic
  segmentation by iterative affinity learning,'' \emph{International Journal of
  Computer Vision}, vol. 128, pp. 1736--1749, 2020.

\bibitem{mccloskey1989catastrophic}
M.~McCloskey and N.~J. Cohen, ``Catastrophic interference in connectionist
  networks: The sequential learning problem,'' in \emph{Psychology of Learning
  and Motivation}, 1989, vol.~24, pp. 109--165.

\bibitem{Goodfellow14anempirical}
I.~J. Goodfellow, M.~Mirza, D.~Xiao, A.~Courville, and Y.~Bengio, ``An
  empirical investigation of catastrophic forgeting in gradientbased neural
  networks,'' in \emph{In Proceedings of International Conference on Learning
  Representations}, 2014.

\bibitem{pfulb2019comprehensive}
B.~Pf{\"u}lb and A.~Gepperth, ``A comprehensive, application-oriented study of
  catastrophic forgetting in dnns,'' in \emph{In Proceedings of International
  Conference on Learning Representations}, 2019.

\bibitem{continualReview19}
G.~I. Parisi, R.~Kemker, J.~L. Part, C.~Kanan, and S.~Wermter, ``Continual
  lifelong learning with neural networks: A review,'' \emph{Neural Networks},
  vol. 113, pp. 54--71, 2019.

\bibitem{lomonaco2017core50}
V.~Lomonaco and D.~Maltoni, ``Core50: a new dataset and benchmark for
  continuous object recognition,'' in \emph{Proceedings of the First Annual
  Conference on Robot Learning}, vol.~78, 2017, pp. 17--26.

\bibitem{camoriano2017incremental}
R.~Camoriano, G.~Pasquale, C.~Ciliberto, L.~Natale, L.~Rosasco, and G.~Metta,
  ``Incremental robot learning of new objects with fixed update time,'' in
  \emph{2017 IEEE International Conference on Robotics and Automation}, 2017,
  pp. 3207--3214.

\bibitem{de2019continual}
M.~De~Lange, R.~Aljundi, M.~Masana, S.~Parisot, X.~Jia, A.~Leonardis,
  G.~Slabaugh, and T.~Tuytelaars, ``Continual learning: A comparative study on
  how to defy forgetting in classification tasks,'' \emph{arXiv preprint
  arXiv:1909.08383}, vol.~2, no.~6, 2019.

\bibitem{BTCA2015}
B.~Landman, Z.~Xu, J.~Igelsias, M.~Styner, T.~Langerak, and A.~Klein,
  ``Multi-atlas labeling beyond the cranial vault-workshop and challenge,''
  2015.

\bibitem{jimenez2016cloud}
O.~Jimenez-del Toro, H.~M{\"u}ller, M.~Krenn, K.~Gruenberg, A.~A. Taha,
  M.~Winterstein, I.~Eggel, A.~Foncubierta-Rodr{\'\i}guez, O.~Goksel, A.~Jakab
  \emph{et~al.}, ``Cloud-based evaluation of anatomical structure segmentation
  and landmark detection algorithms: Visceral anatomy benchmarks,'' \emph{IEEE
  Transactions on Medical Imaging}, vol.~35, no.~11, pp. 2459--2475, 2016.

\bibitem{kavur2020chaos}
A.~E. Kavur, N.~S. Gezer, M.~Bar{\i}{\c{s}}, P.-H. Conze, V.~Groza, D.~D. Pham,
  S.~Chatterjee, P.~Ernst, S.~{\"O}zkan, B.~Baydar \emph{et~al.}, ``Chaos
  challenge--combined (ct-mr) healthy abdominal organ segmentation,''
  \emph{arXiv preprint arXiv:2001.06535}, 2020.

\bibitem{KiTSDataset}
N.~Heller, S.~McSweeney, M.~T. Peterson, S.~Peterson, J.~Rickman, B.~Stai,
  R.~Tejpaul, M.~Oestreich, P.~Blake, J.~Rosenberg \emph{et~al.}, ``An
  international challenge to use artificial intelligence to define the
  state-of-the-art in kidney and kidney tumor segmentation in ct imaging.''
  \emph{American Society of Clinical Oncology}, vol.~38, no.~6, pp. 626--626,
  2020.

\bibitem{rister2020ct}
B.~Rister, D.~Yi, K.~Shivakumar, T.~Nobashi, and D.~L. Rubin, ``Ct-org, a new
  dataset for multiple organ segmentation in computed tomography,''
  \emph{Scientific Data}, vol.~7, no.~1, pp. 1--9, 2020.

\bibitem{ronneberger20163DUNet}
{\"O}.~{\c{C}}i{\c{c}}ek, A.~Abdulkadir, S.~S. Lienkamp, T.~Brox, and
  O.~Ronneberger, ``3d u-net: learning dense volumetric segmentation from
  sparse annotation,'' in \emph{International Conference on Medical Image
  Computing and Computer-assisted Intervention}, 2016, pp. 424--432.

\bibitem{ma2021SOTA-Seg}
J.~Ma, ``Cutting-edge 3d medical image segmentation methods in 2020: Are happy
  families all alike?'' \emph{arXiv preprint arXiv:2101.00232}, 2021.

\bibitem{milletari2016Dice}
F.~Milletari, N.~Navab, and S.-A. Ahmadi, ``V-net: Fully convolutional neural
  networks for volumetric medical image segmentation,'' in \emph{2016 Fourth
  International Conference on 3D vision}, 2016, pp. 565--571.

\bibitem{SegLossOdyssey}
J.~Ma, J.~Chen, M.~Ng, R.~Huang, Y.~Li, C.~Li, X.~Yang, and A.~Martel, ``Loss
  odyssey in medical image segmentation,'' \emph{Medical Image Analysis},
  vol.~71, p. 102035, 2021.

\bibitem{perazzi2016DAVIS}
F.~Perazzi, J.~Pont-Tuset, B.~McWilliams, L.~Van~Gool, M.~Gross, and
  A.~Sorkine-Hornung, ``A benchmark dataset and evaluation methodology for
  video object segmentation,'' in \emph{Proceedings of the IEEE Conference on
  Computer Vision and Pattern Recognition}, 2016, pp. 724--732.

\bibitem{pont2017DAVIS}
J.~Pont-Tuset, F.~Perazzi, S.~Caelles, P.~Arbel{\'a}ez, A.~Sorkine-Hornung, and
  L.~Van~Gool, ``The 2017 davis challenge on video object segmentation,''
  \emph{arXiv preprint arXiv:1704.00675}, 2017.

\bibitem{CVPR20-noisy-student}
Q.~Xie, M.-T. Luong, E.~Hovy, and Q.~V. Le, ``Self-training with noisy student
  improves imagenet classification,'' in \emph{Proceedings of the IEEE/CVF
  Conference on Computer Vision and Pattern Recognition}, 2020, pp.
  10\,687--10\,698.

\bibitem{ECCV20-noisy-student-seg}
L.-C. Chen, R.~G. Lopes, B.~Cheng, M.~D. Collins, E.~D. Cubuk, B.~Zoph,
  H.~Adam, and J.~Shlens, ``Semi-supervised learning in video sequences for
  urban scene segmentation,'' \emph{European Conference on Computer Vision},
  2020.

\bibitem{zhang2019sparse}
Z.~Zhang, J.~Li, Z.~Zhong, Z.~Jiao, and X.~Gao, ``A sparse annotation strategy
  based on attention-guided active learning for 3d medical image
  segmentation,'' \emph{arXiv preprint arXiv:1906.07367}, 2019.

\bibitem{FullyCRFs}
P.~Kr{\"a}henb{\"u}hl and V.~Koltun, ``Efficient inference in fully connected
  crfs with gaussian edge potentials,'' in \emph{Advances in Neural Information
  Processing Systems}, vol.~24, 2011, pp. 109--117.

\bibitem{gao2016CNN-CRF}
M.~Gao, Z.~Xu, L.~Lu, A.~Wu, I.~Nogues, R.~M. Summers, and D.~J. Mollura,
  ``Segmentation label propagation using deep convolutional neural networks and
  dense conditional random field,'' in \emph{2016 IEEE 13th International
  Symposium on Biomedical Imaging}, 2016, pp. 1265--1268.

\bibitem{MICCAI16-CNN-CRF-Liver}
P.~F. Christ, M.~E.~A. Elshaer, F.~Ettlinger, S.~Tatavarty, M.~Bickel,
  P.~Bilic, M.~Rempfler, M.~Armbruster, F.~Hofmann, M.~D’Anastasi
  \emph{et~al.}, ``Automatic liver and lesion segmentation in ct using cascaded
  fully convolutional neural networks and 3d conditional random fields,'' in
  \emph{International Conference on Medical Image Computing and
  Computer-Assisted Intervention}, 2016, pp. 415--423.

\bibitem{Yao17-CRF-Liver}
Y.~{Zhang}, Z.~{He}, C.~{Zhong}, Y.~{Zhang}, and Z.~{Shi}, ``Fully
  convolutional neural network with post-processing methods for automatic liver
  segmentation from ct,'' in \emph{2017 Chinese Automation Congress}, 2017, pp.
  3864--3869.

\bibitem{MIA18-CNN-CRF-BrainTumor}
X.~Zhao, Y.~Wu, G.~Song, Z.~Li, Y.~Zhang, and Y.~Fan, ``A deep learning model
  integrating fcnns and crfs for brain tumor segmentation,'' \emph{Medical
  Image Analysis}, vol.~43, pp. 98--111, 2018.

\bibitem{BraTS18-1st}
A.~Myronenko, ``3d mri brain tumor segmentation using autoencoder
  regularization,'' in \emph{International MICCAI Brainlesion Workshop}, 2018,
  pp. 311--320.

\bibitem{TPAMI17-LearnNoForget}
Z.~Li and D.~Hoiem, ``Learning without forgetting,'' \emph{IEEE Transactions on
  Pattern Analysis and Machine Intelligence}, vol.~40, no.~12, pp. 2935--2947,
  2017.

\bibitem{xie2019TinyOrgan}
L.~Xie, Q.~Yu, Y.~Zhou, Y.~Wang, E.~K. Fishman, and A.~L. Yuille, ``Recurrent
  saliency transformation network for tiny target segmentation in abdominal ct
  scans,'' \emph{IEEE Transactions on Medical Imaging}, vol.~39, no.~2, pp.
  514--525, 2019.

\end{thebibliography}
\end{document}